\documentclass[journal]{IEEEtran}
\IEEEoverridecommandlockouts                              
\usepackage{graphicx}
\usepackage{cite}
\usepackage{picinpar}
\usepackage{amsmath}
\usepackage{url}
\usepackage{flushend}
\usepackage[latin1]{inputenc}
\usepackage{colortbl}
\usepackage{soul}
\usepackage{multirow}
\usepackage{pifont}
\usepackage{color}
\usepackage{alltt}
\usepackage[hidelinks]{hyperref}
\usepackage{enumerate}
\usepackage{siunitx}
\usepackage{breakurl}
\usepackage{epstopdf}
\usepackage{pbox}
\usepackage{subfig}
\usepackage[linesnumbered,ruled,vlined]{algorithm2e}
\usepackage{booktabs}
\usepackage[table,xcdraw]{xcolor}

\begin{document}
\newtheorem{assumption}{Assumption}
\title{Simulation-to-reality UAV Fault Diagnosis in windy environments}
%
%
%

\author{Wei Zhang, Junjie Tong, Fang Liao, Yunfeng Zhang
\thanks{Wei Zhang, Junjie Tong and Yunfeng Zhang are with the Department of Mechanical Engineering, National University of Singapore, Singapore 117575, Singapore.
	{\tt\small e-mail: weizhang@u.nus.edu, e0240715@u.nus.edu, mpezyf@nus.edu.sg}
	}%
\thanks{Fang Liao is with Temasek Laboratories, National University of Singapore, Singapore 117411, Singapore.
	{\tt\small e-mail:tsllf@nus.edu.sg}
	}%
}
\markboth{}
{Wei Zhang \MakeLowercase{\textit{et al.}}: Simulation-to-reality UAV Fault Diagnosis in windy environments} 
\maketitle
\begin{abstract}
Monitoring propeller failures is vital to maintain the safe and reliable operation of quadrotor UAVs. The simulation-to-reality UAV fault diagnosis technique offer a secure and economical approach to identify faults in propellers. However, classifiers trained with simulated data perform poorly in real flights due to the wind disturbance in outdoor scenarios. In this work, we propose an uncertainty-based fault classifier (UFC) to address the challenge of sim-to-real UAV fault diagnosis in windy scenarios. It uses the ensemble of difference-based deep convolutional neural networks (EDDCNN) to reduce model variance and bias. Moreover, it employs an uncertainty-based decision framework to filter out uncertain predictions. Experimental results demonstrate that the UFC can achieve 100\% fault-diagnosis accuracy with a data usage rate of 33.6\% in the windy outdoor scenario.
\end{abstract}


%
\IEEEpeerreviewmaketitle
\section{Introduction}

\begin{figure}[htp]
	\centering
	\includegraphics[width=0.8\linewidth]{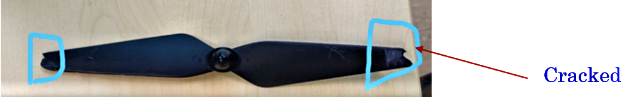}
	\caption{Example of broken propeller \cite{zhang2023D}.}
	\label{propeller}
\end{figure}

\IEEEPARstart{T}{he} utilization of quadrotor unmanned aerial vehicles (UAVs) has gained popularity in diverse domains, including search-and-rescue operations and homeland security \cite{shraim2018survey}. However, UAV propellers are susceptible to damage while in operation (refer to Fig. \ref{propeller}), primarily due to unexpected occurrences such as collisions with buildings or trees. Such damage poses a threat to the successful completion of the UAV's mission and may result in significant harm to both the UAV itself and the surrounding individuals. Hence, regular monitoring of the quadrotor propellers is imperative. Early identification of propeller malfunctions enables prompt UAV recall, thus mitigating further damage.

Recently, data-driven approaches have gained considerable attention in the field of UAV fault diagnosis \cite{iannace2019fault}, \cite{park2022multiclass}. In contrast to model-based methods \cite{falcon2022robust}, data-driven techniques do not depend on a system model. In these approaches, deep neural networks (DNNs) served as the fault classifier and were trained using flight data \cite{hinton2006reducing}. The DNN fault classifiers operate in an end-to-end fashion and do not require additional feature extraction. Compared to other areas of data-driven fault diagnosis, such as machinery fault diagnosis \cite{Zhang2018}, obtaining data on UAV faults is challenging due to the potential risk of drone crashes resulting from flight with a faulty propeller. To mitigate the risk of collecting data in real flight, training the real-world fault classifier with simulation data is much safer and cheaper. This idea, referred to as simulation-to-reality (sim-to-real), is widely used for robot control \cite{Jemin2019}, \cite{zhang2022ipaprec}. Recently, the sim-to-real idea has been adopted in fault diagnosis domain, such as bearing fault diagnosis \cite{Xiao2022}. Zhang et al. \cite{zhang2023D} presented the difference-based deep convolutional neural network (DDCNN), which successfully addressed the sim-to-real UAV fault diagnosis problem in indoor scenarios. However, when conducting outdoor missions, UAVs are subject to wind disturbances that can significantly impact their performance. The effects of wind on UAVs also pose a significant challenge to fault classifiers.

The aim of this research is to develop an accurate sim-to-real fault classifier capable of detecting UAV faults in outdoor windy environments. In order to achieve this objective, we present the uncertainty-based fault classifier (UFC) to filter out uncertain predictions and retain highly confident ones, which in turn enhances the accuracy and reliability of the diagnostic results. In addition, DDCNN ensemble (EDDCNN) is presented as the low-level fault classifier under the framework of UFC. It can mitigate the variance and bias of the DDCNN model. To sum up, the contributions of this study are outlined as follows,
\begin{itemize}
    \item UFC is presented to enhance the confidence of the UAV fault classifier in its predictions. It allows for more accurate and reliable identification of UAV faults in challenging outdoor environments.
	\item The EDDCNN model is presented to reduce the variance and bias of the DDCNN model. It also provides a better estimate of the predictive uncertainty than the single DDCNN model. 
    \item With UFC, the EDDCNN model with ten DDCNN models can achieve 100\% classification accuracy with a data usage of 33.6\%.
\end{itemize}

The rest of this work is structured as follows. A concise overview of the relevant literature is given in Section \ref{RW}, then the problem description is presented in Section \ref{Problem description}. In Section \ref{approach}, we introduce our proposed solution, the EDDCNN model and UFC, describing their respective features and benefits. We then provide a comprehensive performance analysis of our approach, including model training and evaluation, in Section \ref{Implementation}. Lastly, we draw conclusions and give potential future directions in Section \ref{Conclusion}.

\section{Related works}\label{RW}
\subsection{Data-driven UAV fault diagnosis}
Recently, data-driven approaches have show great potential in diagnosing the propeller fault in quadrotor UAVs \cite{Puchalski2022}. Yang et al. \cite{yang2021intelligent} proposed a deep residual shrinkage network to classify the fault of UAV propellers. The trained fault classifier showed good diagnosis accuracy in real flight data. However, the trained data and the testing data were collected in the same scenario, the capability of diagnosing the UAV in different scenarios, such as in different wind conditions, were unknown. Besides, Park et al. \cite{park2022multiclass} presented a method based on stacked pruning sparse denoising auto-encoder for UAV fault detection, which demonstrated good performance in noisy scenarios thanks to the use of the auto-encoder. Li et al. \cite{Li2022} also presented an approach using siamese neural network \cite{Taigman2014} for diagnosing fixed-wing UAVs with limited training samples, but its performance advantage was limited to such scenarios. Its performance became comparable to conventional classifiers like support vector machine \cite{Hearst1998} when the training sample size increased.

An alternative approach to using UAV state information for fault diagnosis tasks is to utilize the audio signals received by an external microphone array. For example, Katta et al. \cite{katta2022real} used convolutional neural networks (CNNs) \cite{krizhevsky2017imagenet} to classify the Mel-frequency cepstral coefficient (MFCC) features of audio signals for fault type identification. However, for the above works, the use of malfunctioning UAV for data collection poses a significant safety hazard due to the possibility of unexpected behavior. Zhang et al. \cite{zhang2023D} proposed DDCNN, which used simulated data as input and achieved high accuracy in indoor real-world scenarios. However, the performance of DDCNN will degrade rapidly in windy outdoor scenarios, which will be given in Section \ref{results}.

\subsection{Uncertainty-based fault diagnosis}
In recent years, uncertainty-based methods have received increasing attention in the field of fault diagnosis.  Han et al. \cite{HAN2022108648} used predictive uncertainty and deep ensemble to help address the out-of-distribution (OOD) problem in machinery fault diagnosis. Once the uncertainty value was above a threshold, the diagnostic system turned to human intervention. However, when the target domain differed a lot from the source domain, such as sim-to-real and wind disturbances, many samples may produce high uncertainty values. Such samples must be checked by human experts, which may make the autonomous diagnostic system human-controlled. Guo et al.\cite{Guo2023} used prediction uncertainty to detect faults in fixed-wing UAV actuators. The uncertainty was estimated using prediction residual calculated by an LSTM-based approximator. However, the fault samples were created from artificially modified healthy flight signals from the same domain, and the cross-domain performance of the method in detecting real faults was unknown. In addition, Zhou et al. \cite{Zhou2023Trustworthy} applied predictive uncertainty to detect unknown faults in bearing fault diagnosis. In the presented decision process, unknown faults were detected when the uncertainty value was high. However, their decision framework did not take into account the influence of external disturbances. For example, for the diagnosis of UAV faults, an all-healthy sample may yield high predictive uncertainty under wind disturbances.

\section{Problem description}\label{Problem description}
The problem of identifying faulty propellers in quadrotors can be approached through a classification framework with five distinct categories, including one for all-healthy and four for various faults, as outlined in Table \ref{labels}. To train the classifier, the fight data of the quadrotor is utilized. The same as \cite{zhang2023}, our paper employs the angular accelerations $(\dot{p},\dot{q},\dot{r})$ and the square of the propeller rotational speed ($\omega^2_{i=1,2,3,4}$) as input at step time $t$, denoted by $x_t$ as follows,
\begin{equation}\label{input_eq}
\begin{aligned}
x_t=\left[\begin{matrix}{\dot{p}}_{t-L}&{\dot{p}}_{t-L+1}&\cdots&{\dot{p}}_t\\{\dot{q}}_{t-L}&{\dot{q}}_{t-L+1}&\cdots&{\dot{q}}_t\\{\dot{r}}_{t-L}&{\dot{r}}_{t-L+1}&\cdots&{\dot{r}}_t\\\omega_{1, t-L}^2&\omega_{1, t-L+1}^2&\cdots&\omega_{1, t}^2\\\omega_{2, t-L}^2&\omega_{2, t-L+1}^2&\cdots&\omega_{2, t}^2\\\omega_{3, t-L}^2&\omega_{3, t-L+1}^2&\cdots&\omega_{3, t}^2\\\omega_{4, t-L}^2&\omega_{4, t-L+1}^2&\cdots&\omega_{4, t}^2\\\end{matrix}\right].
\end{aligned}
\end{equation}
where the variable $L$ represents the length of the time window. The goal of the classifier is to categorize each input sample into one of the five categories listed in Table \ref{labels}, based on their respective outputs.

\begin{table}[htp]
\caption{\label{labels} Labels of five fault categories.}
\centering
\begin{tabular}{llllll}
\hline
\hline
Propeller No. & label 1 & label 2 & label 3 & label 4 & label 5\\ \hline
Propeller 1 & healthy & faulty & healthy & healthy & healthy \\
Propeller 2 & healthy & healthy & faulty & healthy & healthy\\
Propeller 3 & healthy & healthy & healthy & faulty & healthy \\
Propeller 4 & healthy & healthy & healthy & healthy & faulty \\
\hline
\hline
\end{tabular}
\end{table}
This study trains the classifier using simulated samples and all-healthy samples from actual flights, since acquiring samples with faulty propellers from real flights poses risks. After training, the classifier is applied to detect UAV faults in windy outdoor scenarios. This sim-to-real classification problem is essentially a cross-domain classification challenge, where the simulation domain represents the source domain and the real-world domain serves as the target domain. Due to the domain gap induced by installation error and unexpected wind, this cross-domain classification problem is highly challenging. The main objective of our paper is to develop a classifier using simulated data and all-healthy real flight data that can accurately identify UAV faults in outdoor windy scenarios.

\section{APPROACH}\label{approach}
In this section we describe how model uncertainty is calculated  and how to use uncertainty of the prediction to improve the accuracy of fault classifier.

\subsection{Overall framework}\label{OF}
\begin{figure}[t]
	\centering
	\includegraphics[width=1.0\linewidth]{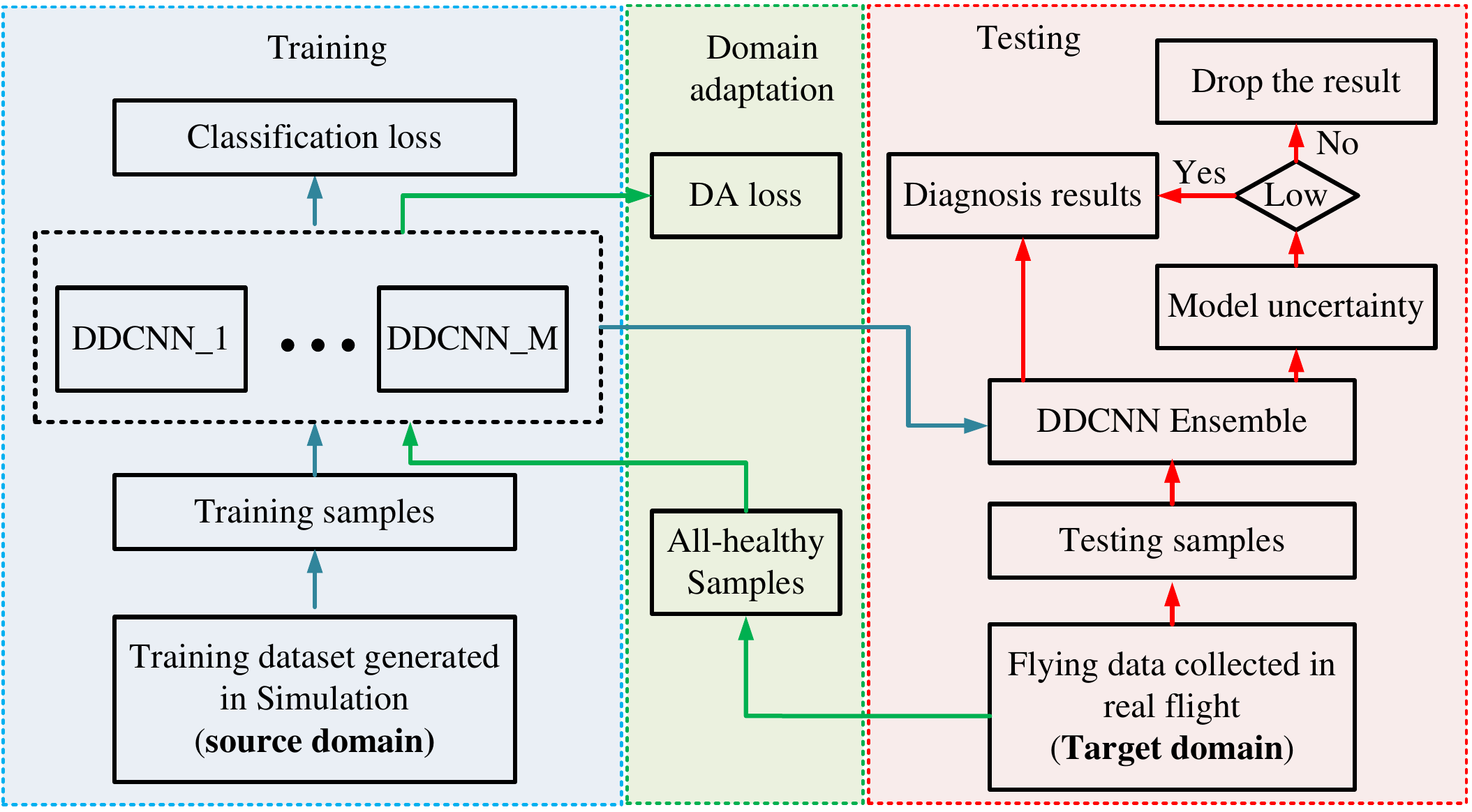}
	\caption{Framework of the proposed approach.}
	\label{Framework}
\end{figure}
The overall framework of the proposed approach is given in Fig. \ref{Framework}. As shown in the training part of the framework, $N$ fault classifiers are trained individually with the simulated training samples to learn the fault classification task. In this paper, difference-based deep convolutional neural network (DDCNN) \cite{zhang2023D} is the utilized as the fault classifier due to its high performance for fault diagnosis in indoor scenarios. During training, each DDCNN is initialized with different weights. The purposes of using multiple classifiers are using model ensemble to reduce training variance and bias and to get a better estimation of the model uncertainty. 


In addition, the same as \cite{zhang2023D} all-healthy samples collected from a single flight before the task serves as the training data for the domain adaptation (DA) task. The objective of DA is to reduce the bring the distributions of all-healthy samples in simulation closer to the counterpart in real flight. The training loss of the DDCNN model is as follows,
\begin{equation}\label{loss}
\begin{aligned}
\mathcal{L} = \mathcal{L}_{c}+\lambda\mathcal{L}_{DA}
\end{aligned}
\end{equation}
where $\mathcal{L}_{c}$ denotes the cross entropy loss for classification; $\mathcal{L}_{DA}$ denotes the DA loss; the factor $\lambda$ is a constant that weighs the contribution of the DA loss. Detailed descriptions of DDCNN can be found in \cite{zhang2023D}.

In the testing phase, the trained $N$ classifiers are combined into classifier ensemble. In this paper, we choose the commonly-used model average method to build the classifier ensemble. When diagnosing a real flight sample, the classifier ensemble can give a prediction with predictive uncertainty. When the uncertainty value is higher than a given threshold, the prediction is considered unreliable and will be discarded; otherwise, the prediction will be selected as the final decision.

\begin{algorithm}[t]
\SetAlgoLined
 Initialize parameters of the $N$ DDCNNs $\theta_1,\theta_2,\ldots, \theta_N$\;
 \textbf{Input}: training Dataset A, all-healthy real-flight Dataset B\;
 Copy the all-healthy category in Dataset A and Dataset B into Dataset D and E\;
 \For{\upshape{model number} $i=1$, 2, \ldots,}
 {\For{\upshape{epoch}$=1$, 2, \ldots, }
 { \For{\upshape{training step}$=1$, 2, \ldots, }
  { Randomly sample $m$ samples from Dataset A, B, D and E into a mini-batch $\mathcal{M}_A$, $\mathcal{M}_B$, $\mathcal{M}_D$ and $\mathcal{M}_E$ \;
  Updating the $\theta_i$ by minimizing total loss $\mathcal{L}$ in (\ref{loss}) using gradient decent\;
  }}}
Combine the $N$ trained DDCNN models into EDDCNN using (\ref{softmax})\;
Calculate the uncertainty threshold $T$ using Dataset B and (\ref{Ada_threshold})\;
\textbf{Output}: EDDCNN with uncertainty threshold $T$\;
\caption{EDDCNN training procedure}
\label{algorithm1}
\end{algorithm}

\begin{algorithm}[t]
\SetAlgoLined
 Input: the input representation $x_t$ for flight data at time step $t$\;
 Feed the sample $x_t$ to EDDCNN\;
 Calculate model uncertainty $H\left[y \mid \mathbf{x_t}\right]$ using (\ref{uncet})\;
\uIf{$H\left[y \mid \mathbf{x}\right]<T$}{
calculate the prediction result $\bar{y_t}$ using (\ref{softmax})\;}
\Else{Drop the result\;}
\caption{Decision process of UFC}
\label{algorithm2}
\end{algorithm}

\subsection{Soft Voting}\label{MA}
Once the $N$ classifiers are trained, they will be combined into the classifier ensemble using the ensemble learning algorithms. As DDCNN is chosen as the fault classifier, the output of the classifier is the probability of each fault calculated by the softmax function. In this paper, soft voting \cite{Zhou2012}, a frequently used ensemble learning method for DNNs, is chosen to combine the class probability outputs of the $N$ classifiers. It averages the probability outputs of the $N$ classifiers and can reduce the variance and bias of the classification results. Using soft voting, the classification label $y$ using $N$ individual DDCNN models is as follows,
\begin{equation}\label{softmax}
\begin{aligned}
\bar{y}=\operatorname*{arg\,max}_j \frac{1}{n}\sum_{k=1}^{N}{\frac{e^{z_{j,k}}}{\sum_{i=1}^{5}e^{z_{i,k}}}}.
\end{aligned}
\end{equation}
where $z_{j,k}$ denotes the logits of the $j$-th output neuron of the $k$-th model. We refer to the DDCNN ensemble as EDDCNN.

\subsection{Uncertainty-based fault classifier}\label{MA}

Due to the external disturbance, such as wind, the fault classifier may not have confidence in its output. Different from machinery fault diagnosis, the UAV fault diagnosis results must be reliable enough, i.e., ensuring 100\% accuracy. For example, incorrectly classifying the all-healthy status into fault condition can lead to false recall of a working UAV and can result in mission failure. Therefore, for UAV fault diagnosis, it is more important to ensure the reliability of the results of each diagnosis than to improving the accuracy of the classifier for all test samples. Predictive uncertainty is a widely used metric to measure the confidence of the classifier. When the predictive uncertainty is high, the classification result are not reliable.  With predictive uncertainty, in this paper, we design uncertainty-based rules to filter out unconfident predictions and select highly confident results as the decision of fault diagnosis. We refer to such a classifier as an uncertainty-based fault classifier (UFC).

In this paper, predictive entropy is chosen to measure the uncertainty of the model because of its foundations in information theory \cite{Gal2016}. It represents the mean information content present within the predictive distribution:
\begin{equation}\label{uncet}
\begin{aligned}
H\left[y \mid \mathbf{x}\right]:=-\sum_{c=1}^{5} p\left(y=c \mid \mathbf{x}\right) \log p\left(y=c \mid \mathbf{x}\right).
\end{aligned}
\end{equation}
where $p\left(y=c \mid \mathbf{x}\right)$ denotes the probability of $c$-th category on test sample $x$. As shown in (\ref{uncet}), the maximum value of the predictive entropy is achieved when all classes are predicted with equal probabilities, which indicates the highest uncertainty of the model. In addition, the minimum value of zero is attained when one class has a probability of one and the rest classes have a probability of zero, which indicates the most confident prediction. As there are $N$ individual DDCNN models, the predictive distribution can be approximated as follows,
\begin{equation}\label{pre_en}
\begin{aligned}
p\left(y=c \mid \mathbf{x}\right) = \frac{1}{N}\sum_{k=1}^{N}{p\left(y=c \mid \mathbf{x}, \theta_k\right)}.
\end{aligned}
\end{equation}
With the prediction results in (\ref{softmax}) and the calculated model uncertainty $H\left[y \mid \mathbf{x}\right]$, the final decision of the fault classifier is as follows,
\begin{equation}\label{threshold}
\begin{aligned}
y_t = 
\begin{cases}
\bar{y_t}, & \text{if}\ H\left[y \mid \mathbf{x}\right]<T,\\
\text{Rejection}, & \text{otherwise}.
\end{cases}
\end{aligned}
\end{equation}
where $T$ denotes the uncertainty threshold. As shown in (\ref{threshold}), when the DDCNN model is confident in its prediction result, i.e., $H\left[y \mid \mathbf{x}\right]<T$,  the prediction will be accepted as the fault detection result, otherwise, the prediction result is considered unreliable and will be rejected. Using UFC, the results used for the final decision do not come from all the evaluated samples, so we need to redefine the accuracy $\alpha$ as follows,
\begin{equation}
    \alpha = \frac{Q_p}{Q_T}
\end{equation}
where $Q_p$ denotes the number of correct predictions; $Q_T$ denotes the total number of predictions chosen by the UFC. $\alpha$ is not sufficient to evaluate the performance of UFC, as a high $\alpha$ can be accompanied by a small $Q_T$, which can significantly reduce the efficiency of fault detection. Hence, we introduce the second metric named data usage rate $\rho_i$, which is used to measure the proportion of data that be used to support the final decision of the fault classifier. It is defined as follows,
\begin{equation}\label{data_usage}
    \rho_i = \frac{Q_{T, i}}{Q_i}
\end{equation}
where $Q_i$ is the total number of samples in fault category $i$; $Q_{T, i}$ is number of samples whose uncertainty values are below the threshold $T$ in fault category $i$. As shown in (\ref{data_usage}), a high data usage rate means more data can be used for fault detection, which allows the fault classifier to diagnose faults earlier. It should be noted that the data usage rate $\rho_1$ for the all-healthy category is not of concern, as failure to detect an all-healthy sample does not result in any action by UAV in the decision framework.

The objective of UFC is to maximize the accuracy and the data usage rate, and the uncertainty threshold $T$ is the key factor to achieve a balance between the metrics. Choosing a small threshold value can lead to more reliable diagnostic results, but at the expense of data usage rate. In this paper, the uncertainty threshold $T$ is calculated as follows,
\begin{equation}\label{Ada_threshold}
\begin{aligned}
T = \max(\arg \max_\tau{\alpha_\tau})
\end{aligned}
\end{equation}
where $\alpha_\tau$ is the testing accuracy conditioned on the threshold $\tau$. The dataset used to calculate $\alpha_\tau$ contains the all-healthy samples collected before the task, which is the same dataset used for domain adaptation. As shown in (\ref{Ada_threshold}), the uncertainty threshold is the highest value that can maximize accuracy, which allows the highest data usage rate with the highest precision. It should be noted that when both $Q_T$ and $Q_p$ are zero, $\alpha_\tau$ is also considered to be 100\%, as this case has the same effect as 100\% accuracy in the other conditions. The EDDCNN training procedure and the UFC decision process can be found in Algorithm \ref{algorithm1} and \ref{algorithm2}, respectively.

\section{Implementation and test}\label{Implementation}
In this section, we introduce the training data and training procedure for the EDDCNN model. After the training was completed, the EDDCNN model was adopted by the UFC, which was tested with real flight data collected from outdoor windy conditions.

\subsection{Dataset Description}
As shown in Table \ref{real_dataset}, three datasets, i.e., Dataset A, B, and C, were generated or collected in this work. Dataset A was generated in simulation using the simulator introduced in \cite{tong2023machine}. To simulate the impact of wind on the drone, we utilized the model presented in \cite{Craig2020}, which correlates wind velocity and direction to the  bluff-body and the induced resistance operating on the UAV.

\begin{figure}[t]
    \subfloat[]{\centering\includegraphics[width=0.45\linewidth]{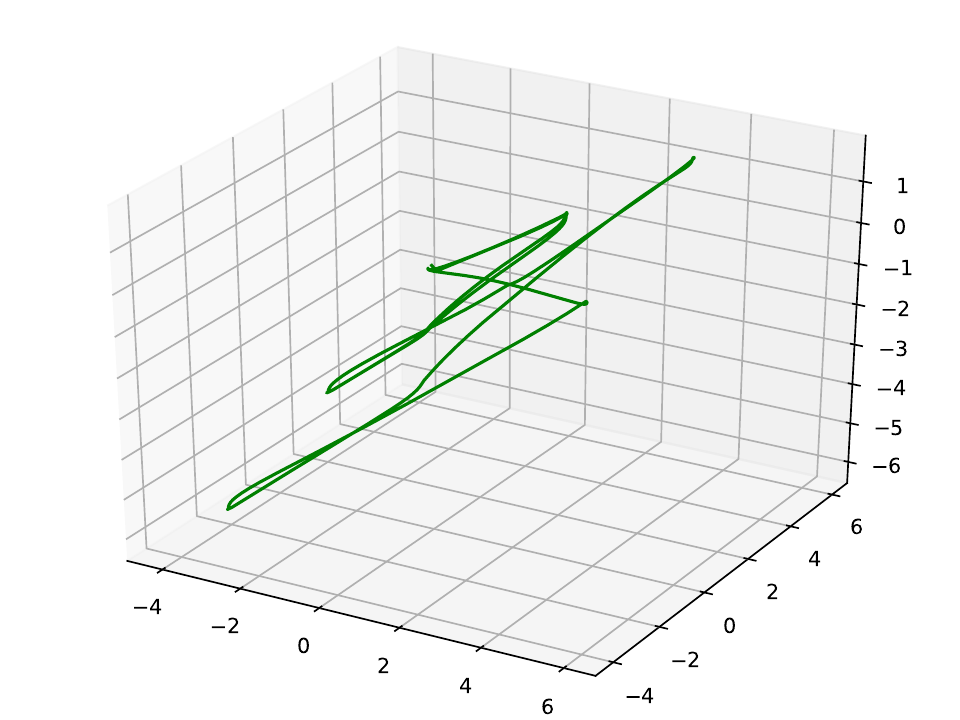}\label{sim_traj}}
    \subfloat[]{\centering\includegraphics[width=0.45\linewidth]{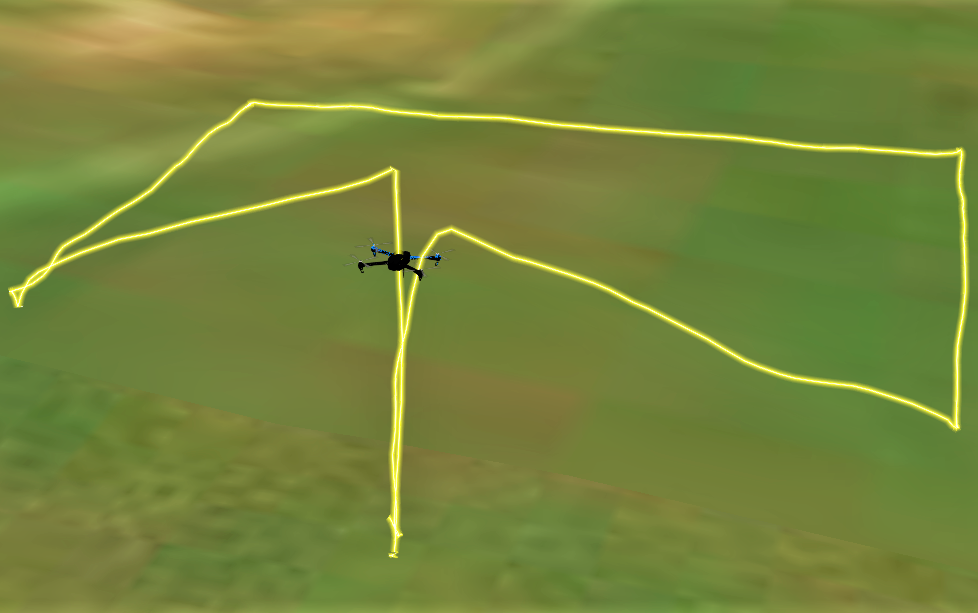}\label{real_traj}}\\
	\subfloat[]{\centering\includegraphics[width=0.9\linewidth]{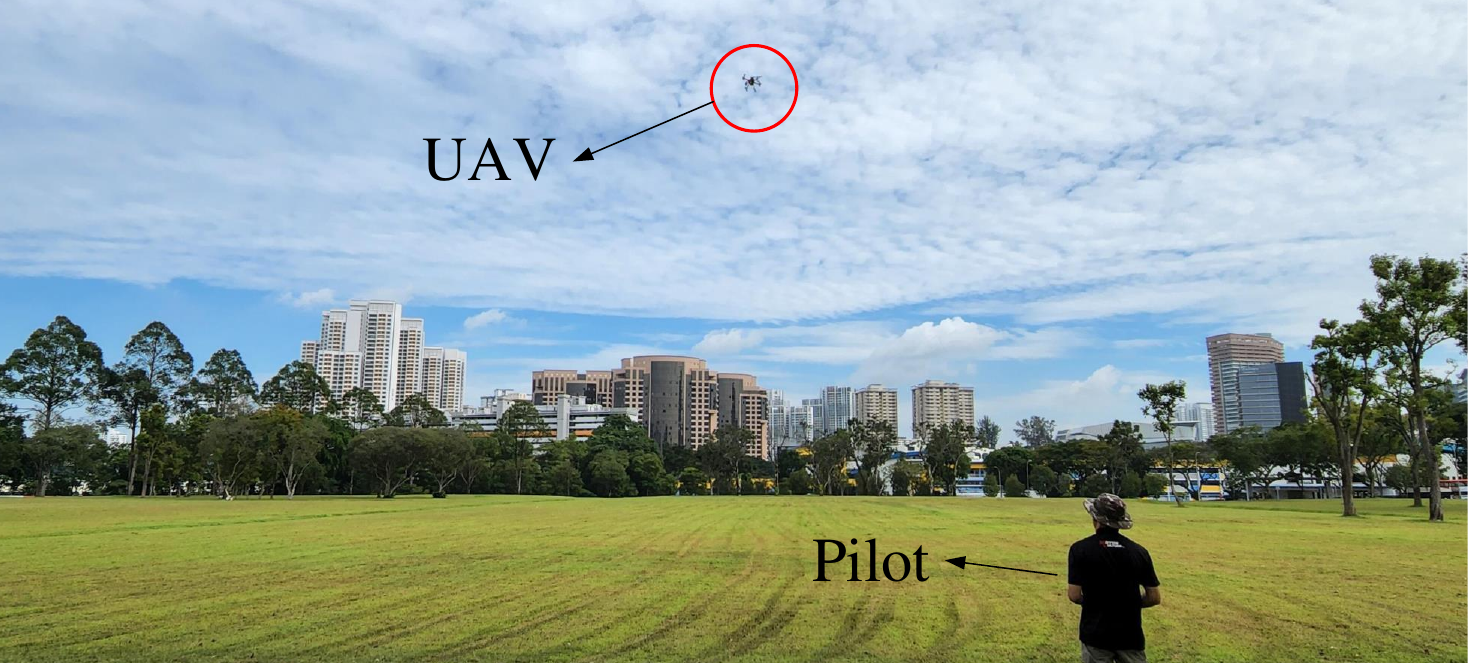}\label{real_exp}}
	\caption{Data collection in simulation and real flight. (a) One of the UAV trajectory for data collection in simulation. (b) One of the UAV trajectories for data collection in real flight. (c) Outdoor scenario for data collection in real flight. }
	\label{data_collection}
\end{figure}

Dataset A consists of 7500 samples divided into five categories, with each category comprising 1500 samples. The samples in each category were produced under three different wind conditions: no wind, 5m/s, and 10m/s wind speeds. In order to collect samples for each fault category, a simulated UAV was employed, which had to fly to specific waypoints using an onboard controller. Figure \ref{sim_traj}
 illustrates one of the UAV's trajectories. Notably, the same waypoints were used for all five categories and wind conditions.

In addition, both training and testing utilized Dataset B, which was obtained from a individual flight conducted before the task. The data collection process required the healthy UAV to navigate to specified waypoints in the outdoor scenario (see Fig. \ref{real_exp}), resulting in the collection of 1500 all-healthy samples. Dataset B served different purposes in training and testing. In the training phase, it was utilized to carry out domain adaptation, while in the testing phase, it aided in computing the average features for the all-healthy category in real-world situations and calculating the threshold of uncertainty.

\begin{figure}[htpb]
    \centering
	  \subfloat[Broken propellers]{
       \includegraphics[width=0.8\linewidth]{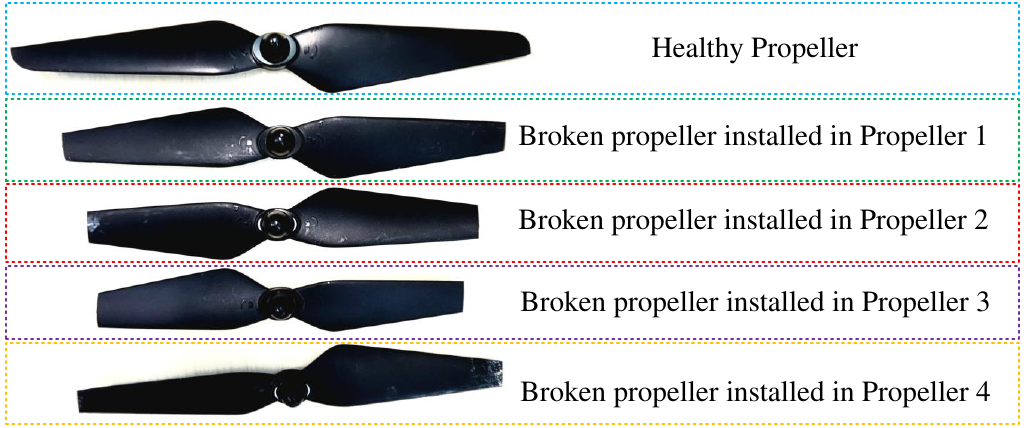}\label{bp}}
        \\
	    \subfloat[Real UAV model]{
        \includegraphics[width=0.9\linewidth]{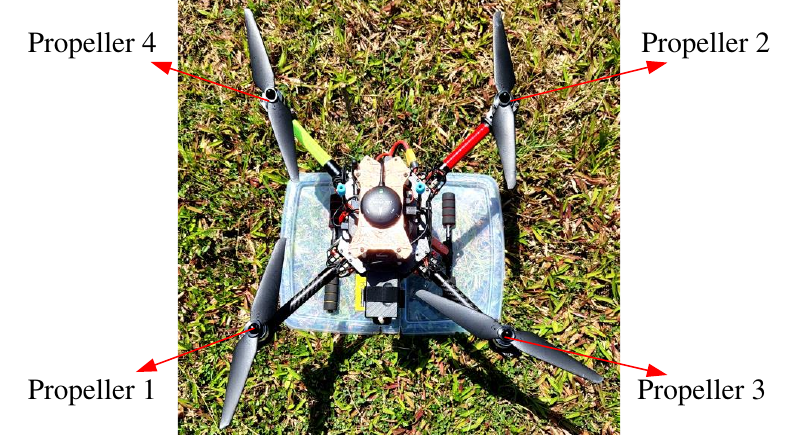}\label{real_UAV}}
	\caption{The real UAV model and broken propellers used in real-world data collection.}
	\label{real_propeller}
\end{figure}

Dataset C, which comprises five categories, was employed for testing purposes. The data was collected from real flights performed in the outdoor scenario shown in Fig. \ref{real_exp}, with 1200, 1200, 1200, 1150 and 900 samples for Label 1 to Label 5, respectively. During flying, the average wind speed was around 3m/s, and the gust speed was around 6m/s. The UAV was required to navigate to designated waypoints during data collection, with a human pilot taking control of the UAV's controller when the onboard controller malfunctioned. One of the flying trajectories is shown in Fig. \ref{real_traj}. Due to unknown reasons, the UAV did not fly to its waypoints for Label 5, and therefore the samples for Label 5 were collected under the control of the human polit, resulting in fewer samples for this label. As shown in Fig. \ref{real_propeller}, Prior to collecting data for each fault category, one healthy propeller was replaced by a broken propeller in the designated position. It is worth noting that the damaged level of the broken propellers in the four locations varied.

\begin{table*}[htp]
\centering
\caption{\label{real_dataset} Description of quadrotor datasets using in training and testing.}
\begin{tabular}{|c|c|c|ccccc|}
\hline
\multirow{3}{*}{Dataset Name} &
  \multirow{3}{*}{Data source} &
  \multirow{3}{*}{Purpose} &
  \multicolumn{5}{c|}{Category label (fault   location)} \\ \cline{4-8} 
 &
   &
   &
  \multicolumn{1}{c|}{1 (None)} &
  \multicolumn{1}{c|}{2 (Propeller 1)} &
  \multicolumn{1}{c|}{3 (Propeller 2)} &
  \multicolumn{1}{c|}{4 (Propeller 3)} &
  5 (propeller 4) \\ \cline{4-8} 
 &
   &
   &
  \multicolumn{5}{c|}{Number of samples} \\ \hline
A &
  Simulation &
  Training &
  \multicolumn{1}{c|}{1500} &
  \multicolumn{1}{c|}{1500} &
  \multicolumn{1}{c|}{1500} &
  \multicolumn{1}{c|}{1500} &
  1500 \\ \hline
B &
  Real flight &
  Training for DA &
  \multicolumn{1}{c|}{1500} &
  \multicolumn{1}{c|}{-} &
  \multicolumn{1}{c|}{-} &
  \multicolumn{1}{c|}{-} & -
   \\ \hline
C &
  Real flight &
  Testing &
  \multicolumn{1}{c|}{1200} &
  \multicolumn{1}{c|}{1200} &
  \multicolumn{1}{c|}{1200} &
  \multicolumn{1}{c|}{1150} &
  900 \\ \hline
\end{tabular}
\end{table*}
\subsection{Training details}
We optimized the DDCNN model using Adam optimizer, setting the learning rate to $3\times10^{-4}$, mini-batch size to 128, and the maximum training epochs to 10. Additionally, we set the Dropout rate to 0.1 and $\lambda=0.05$ for the DA coefficient. The GPU used for training was an RTX 2080Ti, with each model requiring approximately two minutes to complete the training process. Each DDCNN model in the EDDCNN model was trained with different random seed. The uncertainty threshold chosen for the 10-model EDDCNN model was 0.5. The effect of the uncertainty threshold on the different EDDCNN models is examined in section \ref{results}.

\subsection{Results and comparison study}\label{results}
Once trained, the EDDCNN model can make fault diagnosis decisions using the proposed UFC. First, we investigated the effect of the number of models and uncertainty threshold on the testing accuracy. Specifically, we summarized the accuracy of models from 3 to 100 models at uncertainty thresholds ranging from 0.4 to 1.0.  As shown in Table \ref{accuracy_table}, without the support of our uncertainty-based decision framework, the test accuracy was around 60\% for the EDDCNN models with different numbers of base classifiers. With the introduce of the uncertainty threshold, the testing accuracy showed great improvement. When the threshold reached 0.5, the use of 7 models could achieve a testing accuracy of 100\%, which was the ideal accuracy for the diagnosis of UAV faults in the real world.
\begin{table}[]
\footnotesize \setlength{\tabcolsep}{1.6pt}
\renewcommand\arraystretch{1.2}
\caption{\label{accuracy_table} Testing accuracy under different numbers of models in EDDCNN and uncertainty threshold in UFC.}
\begin{tabular}{ccccccccc}
\hline
 & \multicolumn{8}{c}{Uncertainty threshold} \\ \cline{2-9} 
\multirow{-2}{*}{No. of models} & No & 1 & 0.9 & 0.8 & 0.7 & 0.6 & 0.5 & 0.4 \\ \hline
3 & \cellcolor[HTML]{6793CA}59.8\% & \cellcolor[HTML]{7EA3D2}64.7\% & \cellcolor[HTML]{98B6DC}70.1\% & \cellcolor[HTML]{EAEFF8}87.1\% & \cellcolor[HTML]{FCF8FB}91.1\% & \cellcolor[HTML]{FBD4D6}93.3\% & \cellcolor[HTML]{FBB7B9}95.2\% & \cellcolor[HTML]{F8696B}100.0\% \\
5 & \cellcolor[HTML]{6692CA}59.6\% & \cellcolor[HTML]{779ED0}63.2\% & \cellcolor[HTML]{A5BFE0}72.8\% & \cellcolor[HTML]{EAEFF8}87.1\% & \cellcolor[HTML]{FCEBEE}91.9\% & \cellcolor[HTML]{FBC9CC}94.0\% & \cellcolor[HTML]{FA9497}97.3\% & \cellcolor[HTML]{F8696B}100.0\% \\
7 & \cellcolor[HTML]{5A8AC6}57.0\% & \cellcolor[HTML]{749CCF}62.7\% & \cellcolor[HTML]{B4C9E5}75.9\% & \cellcolor[HTML]{F9FAFE}90.3\% & \cellcolor[HTML]{FBD4D7}93.3\% & \cellcolor[HTML]{F98B8D}97.9\% & \cellcolor[HTML]{F8696B}100.0\% & \cellcolor[HTML]{F8696B}100.0\% \\
10 & \cellcolor[HTML]{608EC8}58.3\% & \cellcolor[HTML]{6E98CD}61.3\% & \cellcolor[HTML]{ADC4E3}74.4\% & \cellcolor[HTML]{EEF2FA}87.9\% & \cellcolor[HTML]{FCE0E2}92.6\% & \cellcolor[HTML]{FBB3B6}95.4\% & \cellcolor[HTML]{F8696B}100.0\% & \cellcolor[HTML]{F8696B}100.0\% \\
20 & \cellcolor[HTML]{6893CA}60.0\% & \cellcolor[HTML]{759DCF}62.8\% & \cellcolor[HTML]{98B5DB}70.0\% & \cellcolor[HTML]{F5F7FC}89.4\% & \cellcolor[HTML]{FBC3C5}94.4\% & \cellcolor[HTML]{F8696B}100.0\% & \cellcolor[HTML]{F8696B}100.0\% & \cellcolor[HTML]{F8696B}100.0\% \\
30 & \cellcolor[HTML]{6894CB}60.1\% & \cellcolor[HTML]{83A7D4}65.7\% & \cellcolor[HTML]{B2C8E5}75.5\% & \cellcolor[HTML]{F8F9FD}90.1\% & \cellcolor[HTML]{FBBEC1}94.7\% & \cellcolor[HTML]{F8696B}100.0\% & \cellcolor[HTML]{F8696B}100.0\% & \cellcolor[HTML]{F8696B}100.0\% \\
40 & \cellcolor[HTML]{6894CB}60.1\% & \cellcolor[HTML]{7EA3D2}64.6\% & \cellcolor[HTML]{ADC4E3}74.4\% & \cellcolor[HTML]{F8F9FD}90.0\% & \cellcolor[HTML]{FBC6C8}94.2\% & \cellcolor[HTML]{F8696B}100.0\% & \cellcolor[HTML]{F8696B}100.0\% & \cellcolor[HTML]{F8696B}100.0\% \\
50 & \cellcolor[HTML]{6994CB}60.2\% & \cellcolor[HTML]{82A6D4}65.5\% & \cellcolor[HTML]{AFC6E4}74.9\% & \cellcolor[HTML]{F9F9FD}90.2\% & \cellcolor[HTML]{FBC6C9}94.2\% & \cellcolor[HTML]{F8696B}100.0\% & \cellcolor[HTML]{F8696B}100.0\% & \cellcolor[HTML]{F8696B}100.0\% \\
60 & \cellcolor[HTML]{6692CA}59.7\% & \cellcolor[HTML]{7FA4D3}64.8\% & \cellcolor[HTML]{AFC5E3}74.8\% & \cellcolor[HTML]{F9FAFE}90.3\% & \cellcolor[HTML]{FBBFC2}94.6\% & \cellcolor[HTML]{F8696B}100.0\% & \cellcolor[HTML]{F8696B}100.0\% & \cellcolor[HTML]{F8696B}100.0\% \\
70 & \cellcolor[HTML]{6793CA}59.8\% & \cellcolor[HTML]{81A5D3}65.3\% & \cellcolor[HTML]{BACDE7}77.1\% & \cellcolor[HTML]{FAFBFE}90.5\% & \cellcolor[HTML]{FBC3C5}94.4\% & \cellcolor[HTML]{F8696B}100.0\% & \cellcolor[HTML]{F8696B}100.0\% & \cellcolor[HTML]{F8696B}100.0\% \\
80 & \cellcolor[HTML]{6693CA}59.8\% & \cellcolor[HTML]{7FA4D3}64.9\% & \cellcolor[HTML]{B6CBE6}76.3\% & \cellcolor[HTML]{FAFBFE}90.5\% & \cellcolor[HTML]{FBC1C4}94.5\% & \cellcolor[HTML]{F8696B}100.0\% & \cellcolor[HTML]{F8696B}100.0\% & \cellcolor[HTML]{F8696B}100.0\% \\
90 & \cellcolor[HTML]{6793CA}59.8\% & \cellcolor[HTML]{7FA4D3}64.9\% & \cellcolor[HTML]{B6CBE6}76.4\% & \cellcolor[HTML]{F8F9FD}90.0\% & \cellcolor[HTML]{FBC4C6}94.3\% & \cellcolor[HTML]{F8696B}100.0\% & \cellcolor[HTML]{F8696B}100.0\% & \cellcolor[HTML]{F8696B}100.0\% \\
100 & \cellcolor[HTML]{6693CA}59.7\% & \cellcolor[HTML]{7FA4D3}64.8\% & \cellcolor[HTML]{B9CCE7}76.8\% & \cellcolor[HTML]{F7F9FD}90.0\% & \cellcolor[HTML]{FBC6C9}94.2\% & \cellcolor[HTML]{F8696B}100.0\% & \cellcolor[HTML]{F8696B}100.0\% & \cellcolor[HTML]{F8696B}100.0\% \\ \hline
\end{tabular}
\end{table}

\begin{figure*}[t]
    \centering
	\includegraphics[width=0.8\linewidth]{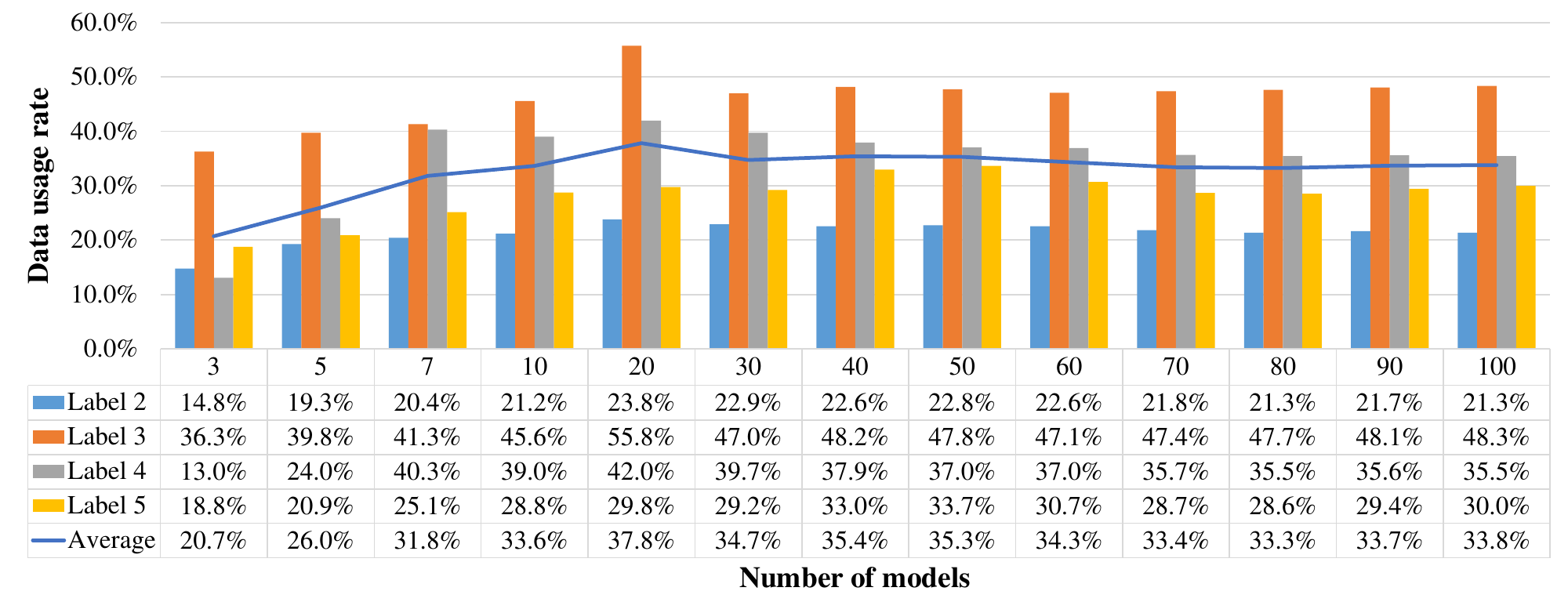}
	\caption{Data usage rate of each fault category for different number of models}
    \label{du}
\end{figure*}

Based on the results presented in Table \ref{accuracy_table}, we conducted further investigation of the proportion of samples that contributed to the 100\% accuracy of UFC. Specifically, we plotted the data usage rate for each fault category with a different number of models in EDDCNN when the UFC first reached its 100\% accuracy. As depicted in Fig. \ref{du}, with 10 models, the average data usage rate reached 33.6\%, indicating that one-third of the total samples could be used to make a final decision. Besides, when there is a higher level of damage to the propeller, such as label 3 and label 4, the corresponding data usage rate is also higher. In addition, the highest data usage rate of 37.8\% was achieved with 20 models. However, increasing the number of models beyond 20 did not increase the data usage rate. To strike a balance between the number of models and the data usage rate, we selected the EDDCNN model that comprised 10 models for the remainder of this study.

\begin{figure}[t]
    \centering
	\includegraphics[width=0.6\linewidth]{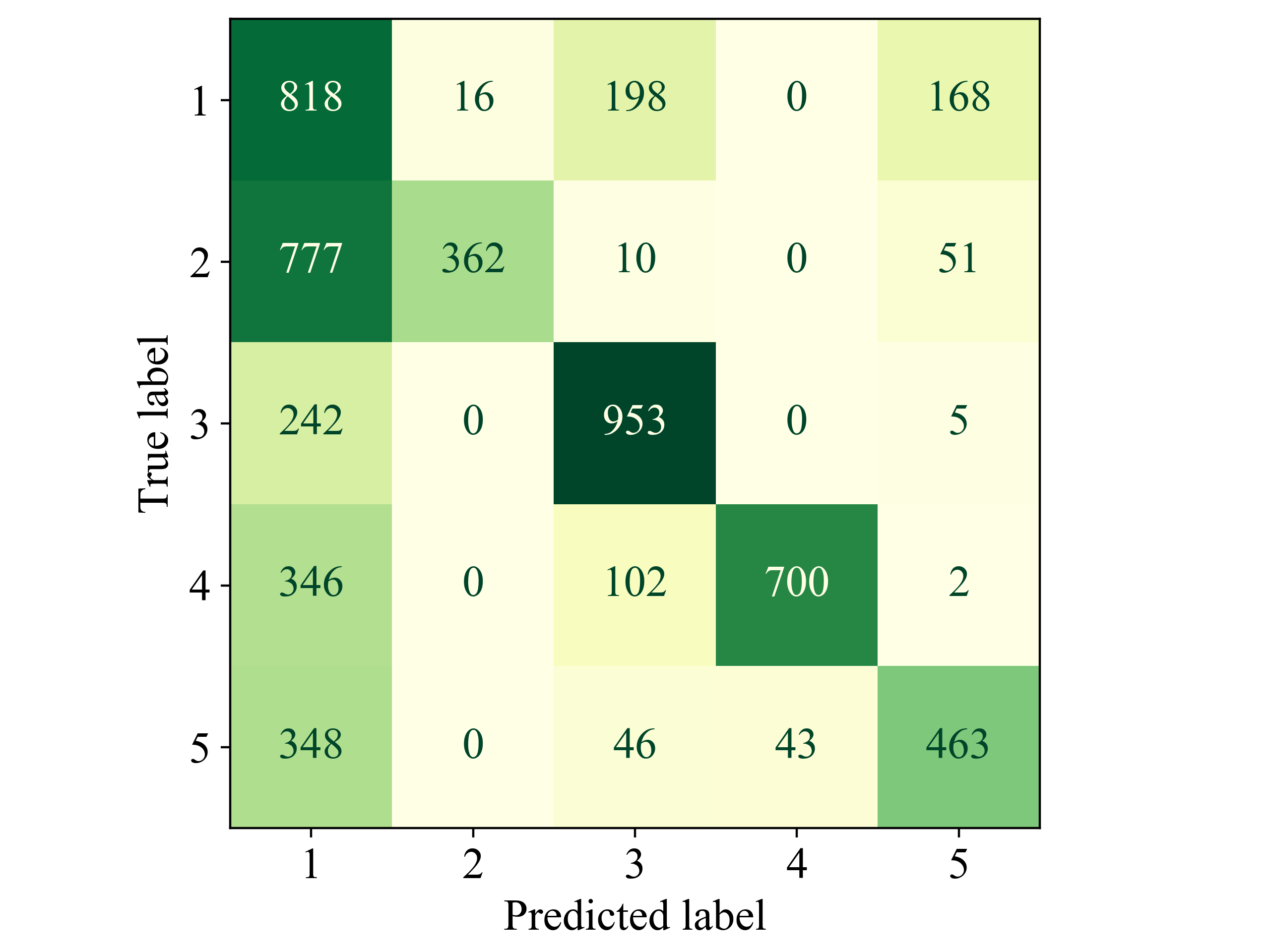}
	\caption{Confusion matrix of the 10-model EDDCNN model without UFC. }
	\label{CM1}
\end{figure}

\begin{figure}
       	\subfloat[$T=1.0$]{\centering\includegraphics[width=0.48\linewidth]{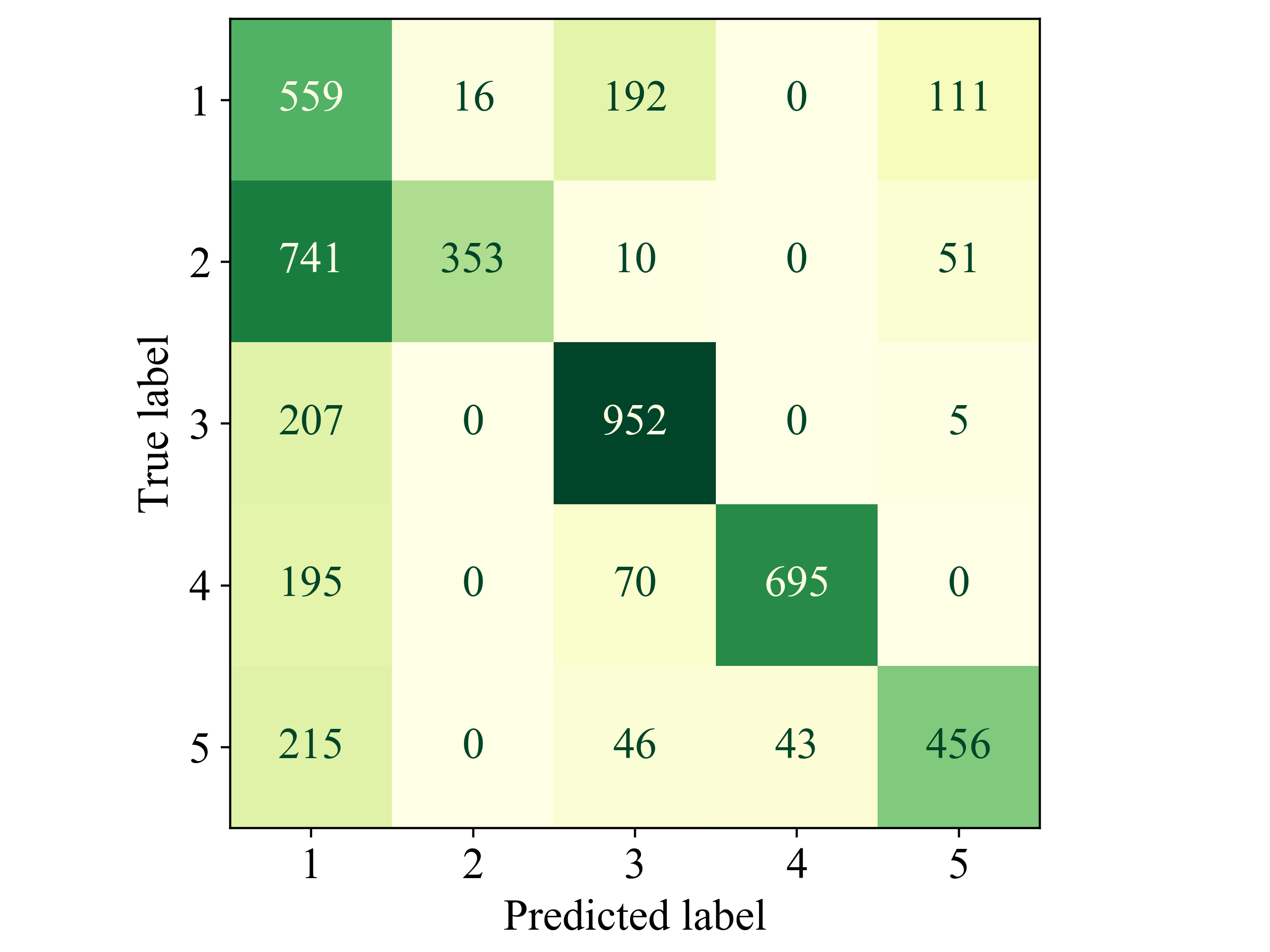}}\label{CM2}
		\subfloat[$T=0.9$]{\centering\includegraphics[width=0.48\linewidth]{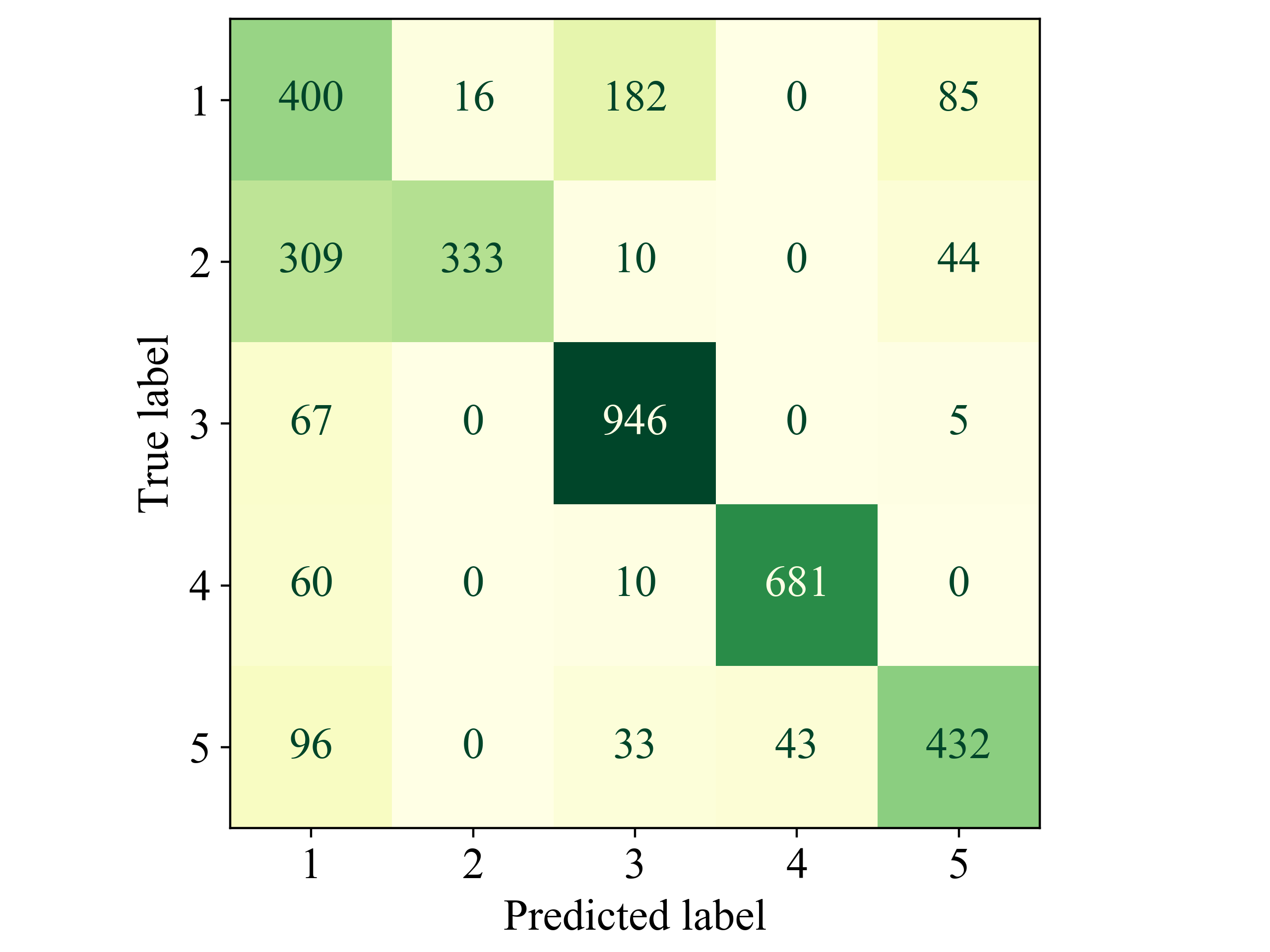}\label{CM3}}\\
       	\subfloat[$T=0.8$]{\centering\includegraphics[width=0.48\linewidth]{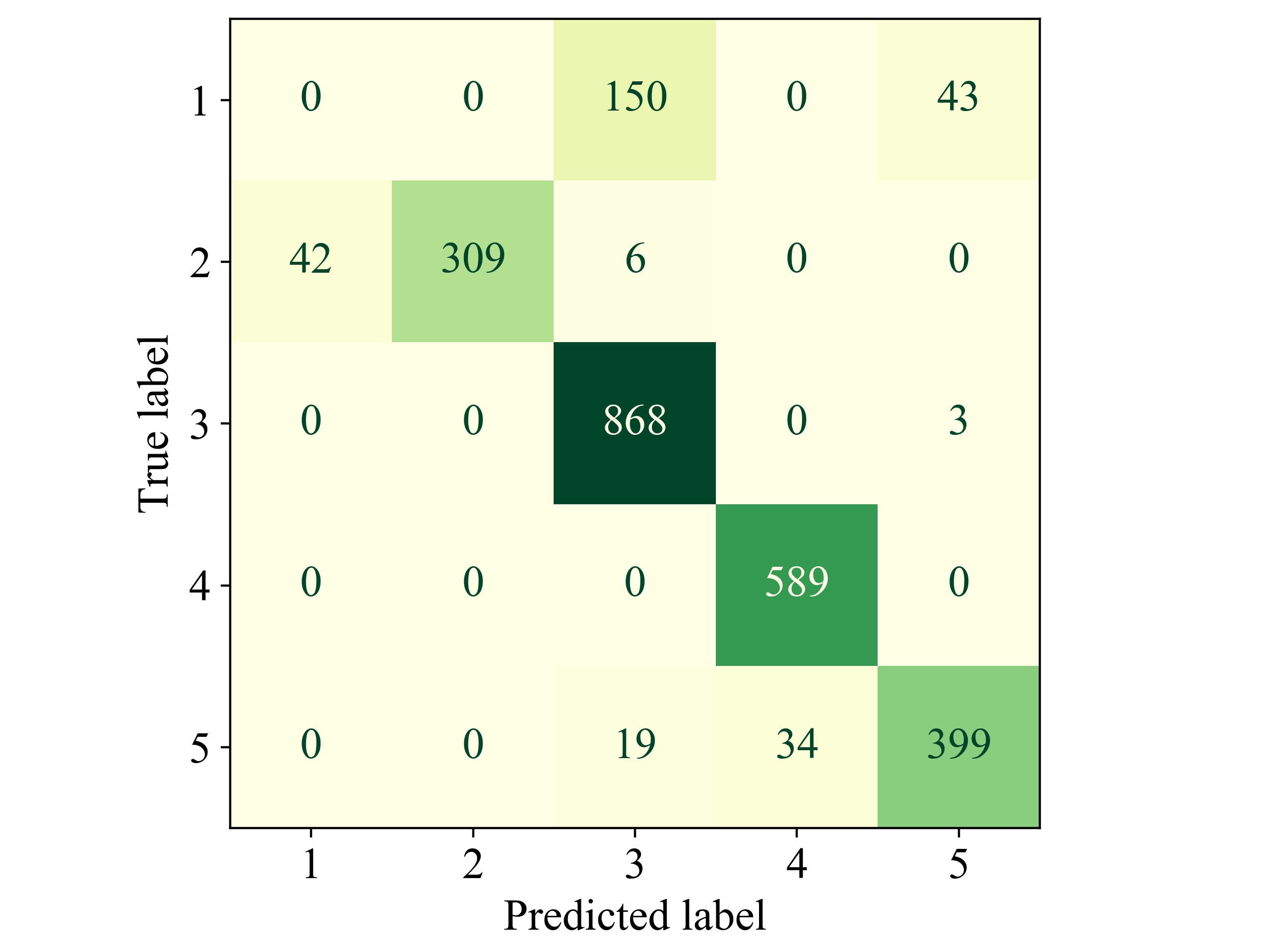}}\label{CM4}
        		\subfloat[$T=0.7$]{\centering\includegraphics[width=0.48\linewidth]{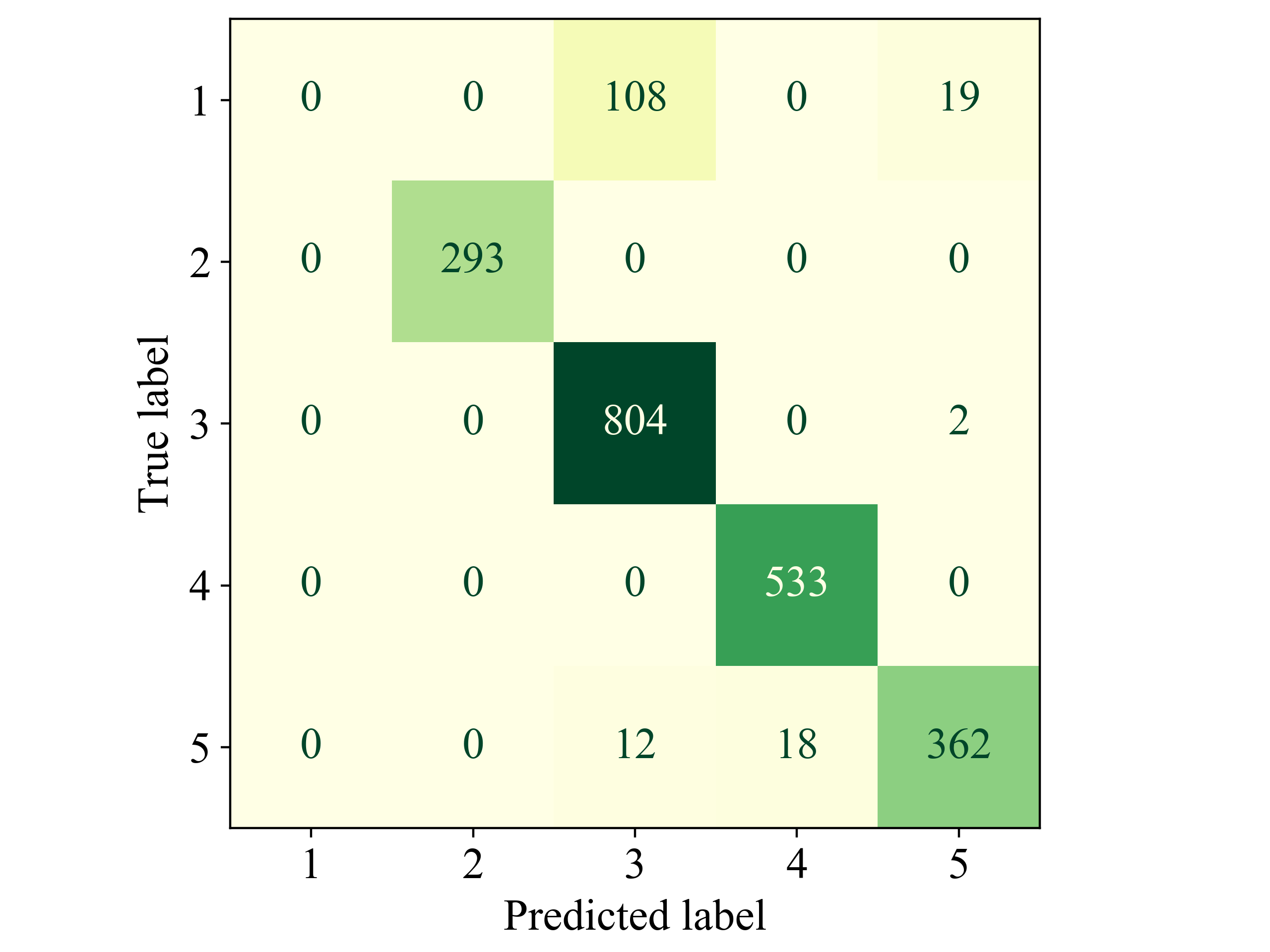}\label{CM5}}\\
       	\subfloat[$T=0.6$]{\centering\includegraphics[width=0.48\linewidth]{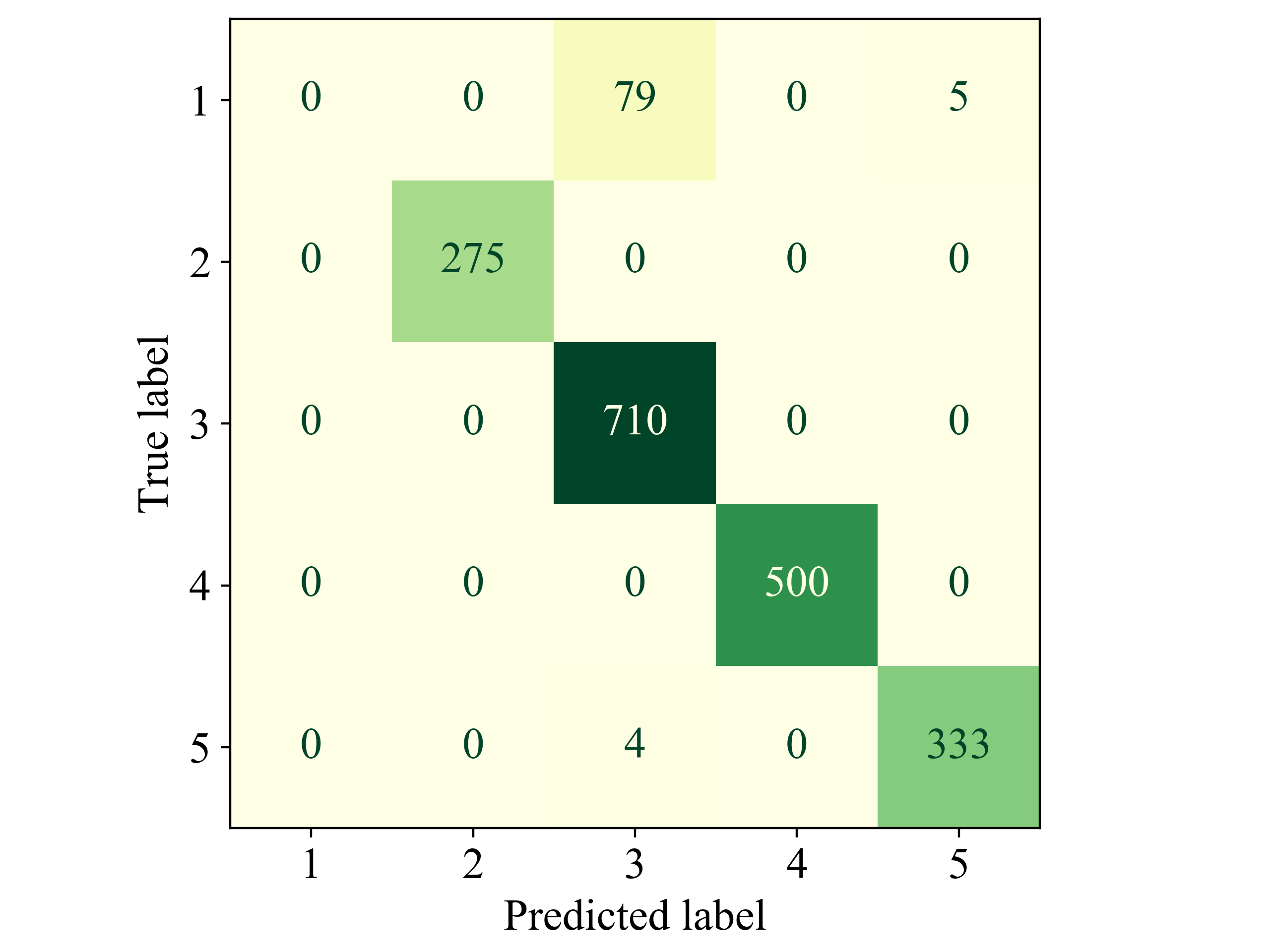}}\label{CM6}
            \subfloat[$T=0.5$]{\centering\includegraphics[width=0.48\linewidth]{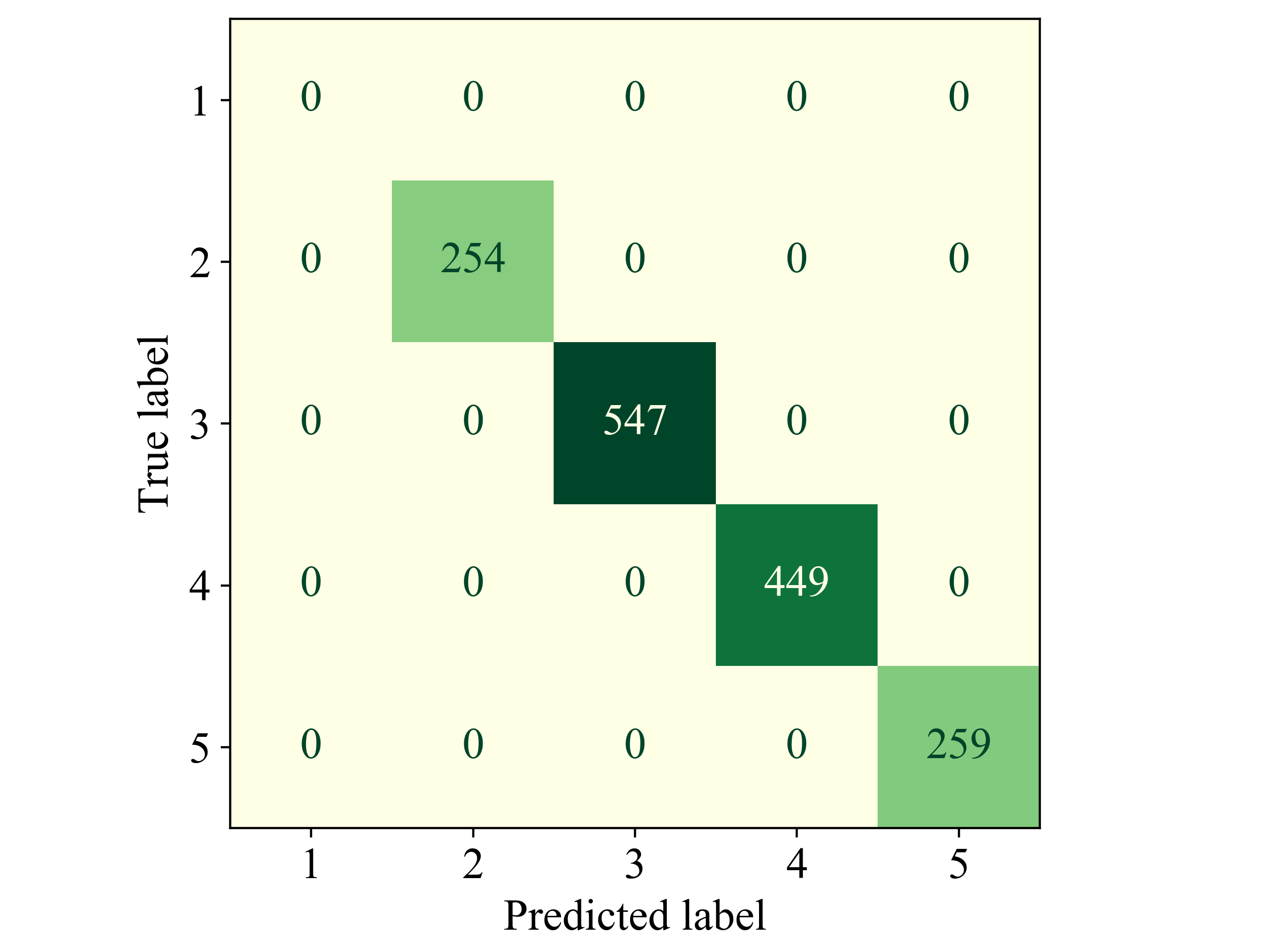}}\label{CM7}
	\caption{Confusion matrix of 10-model EDDCNN model with UFC under different uncertainty threshold.}
	\label{CM_UFC}
\end{figure}

To better understand the UFC filtering process, the confusion matrices without and with UFC are plotted in Fig. \ref{CM1} and Fig. \ref{CM_UFC}, respectively. As depicted in Fig. \ref{CM1}, in the absence of UFC, a substantial number of faulty samples were incorrectly classified into the all-healthy category, while one third of the all-healthy samples were misclassified as faulty. However, as the uncertainty threshold in UFC gradually decreased, the misclassified samples were progressively filtered out. At a threshold value of $T=0.7$, none of the faulty samples were misclassified as all-healthy. Subsequently, at a threshold value of $T=0.5$, all of the all-healthy samples were eliminated by the UFC, and the remaining samples were accurately classified. Although all of the all-healthy samples were ultimately filtered out by the UFC, this did not have any adverse impact on the UAV flight process, since no action was taken.


\subsection{Visualization of the testing result}\label{vis}
\begin{figure}
		\subfloat[Label 1 prediction]{\centering\includegraphics[width=0.4\linewidth]{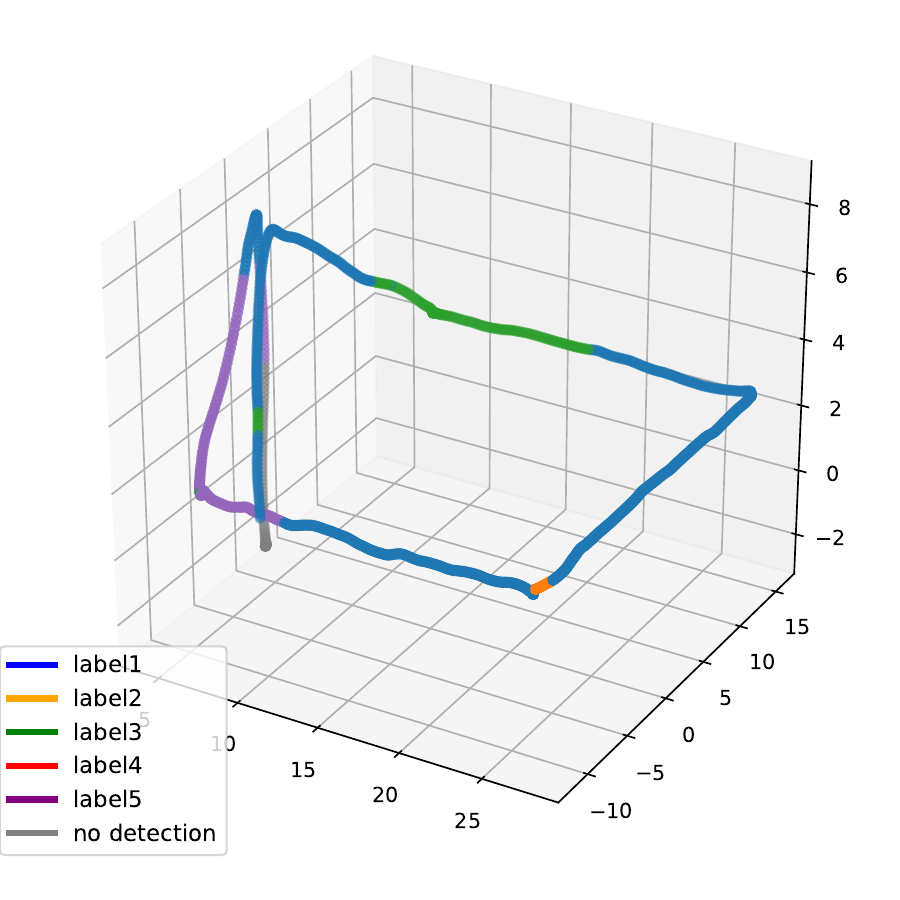}\label{acc1}}
       	\subfloat[Label 1 uncertainty]{\centering\includegraphics[width=0.55\linewidth]{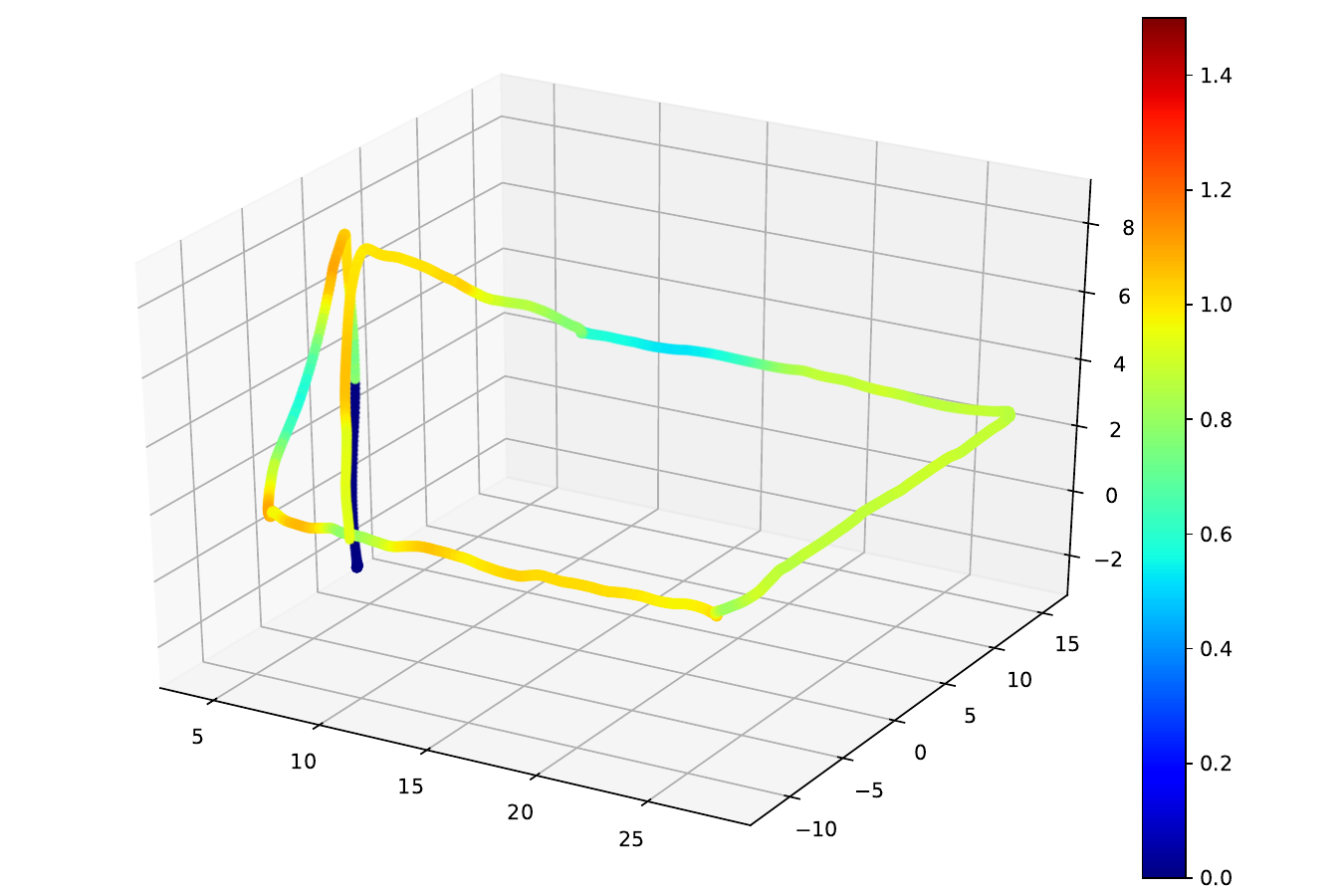}\label{uncertainty1}}\\
		\subfloat[Label 2 prediction]{\centering\includegraphics[width=0.4\linewidth]{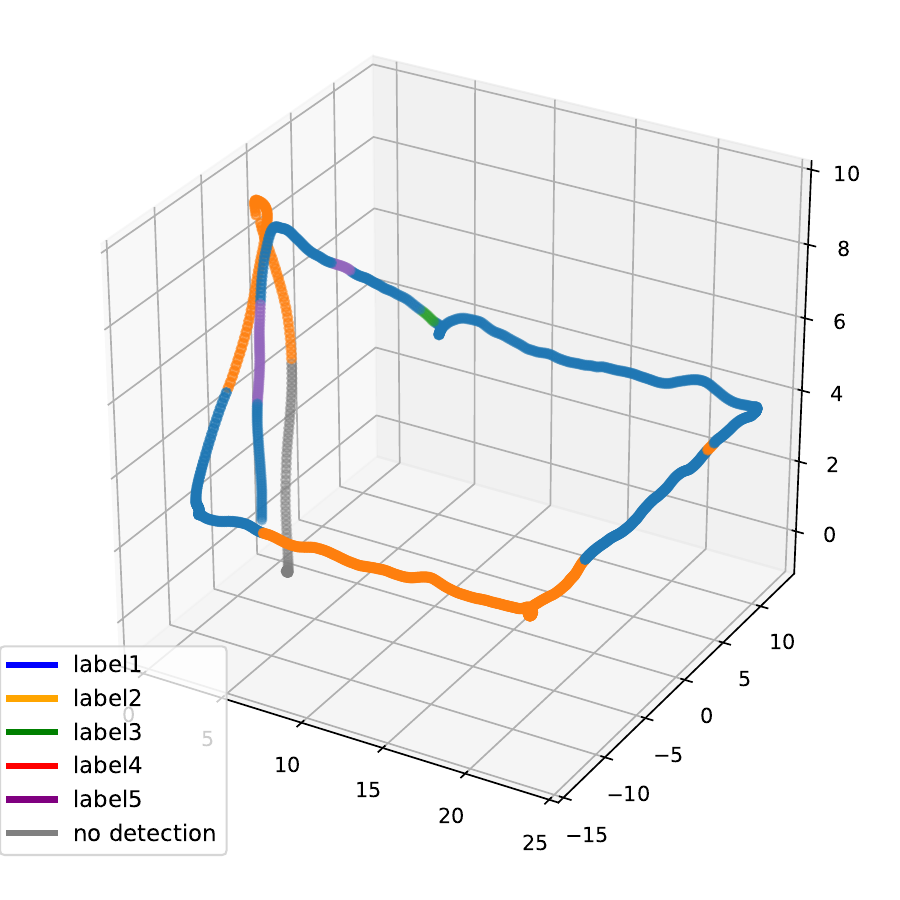}\label{acc2}}
       	\subfloat[Label 2 uncertainty]{\centering\includegraphics[width=0.55\linewidth]{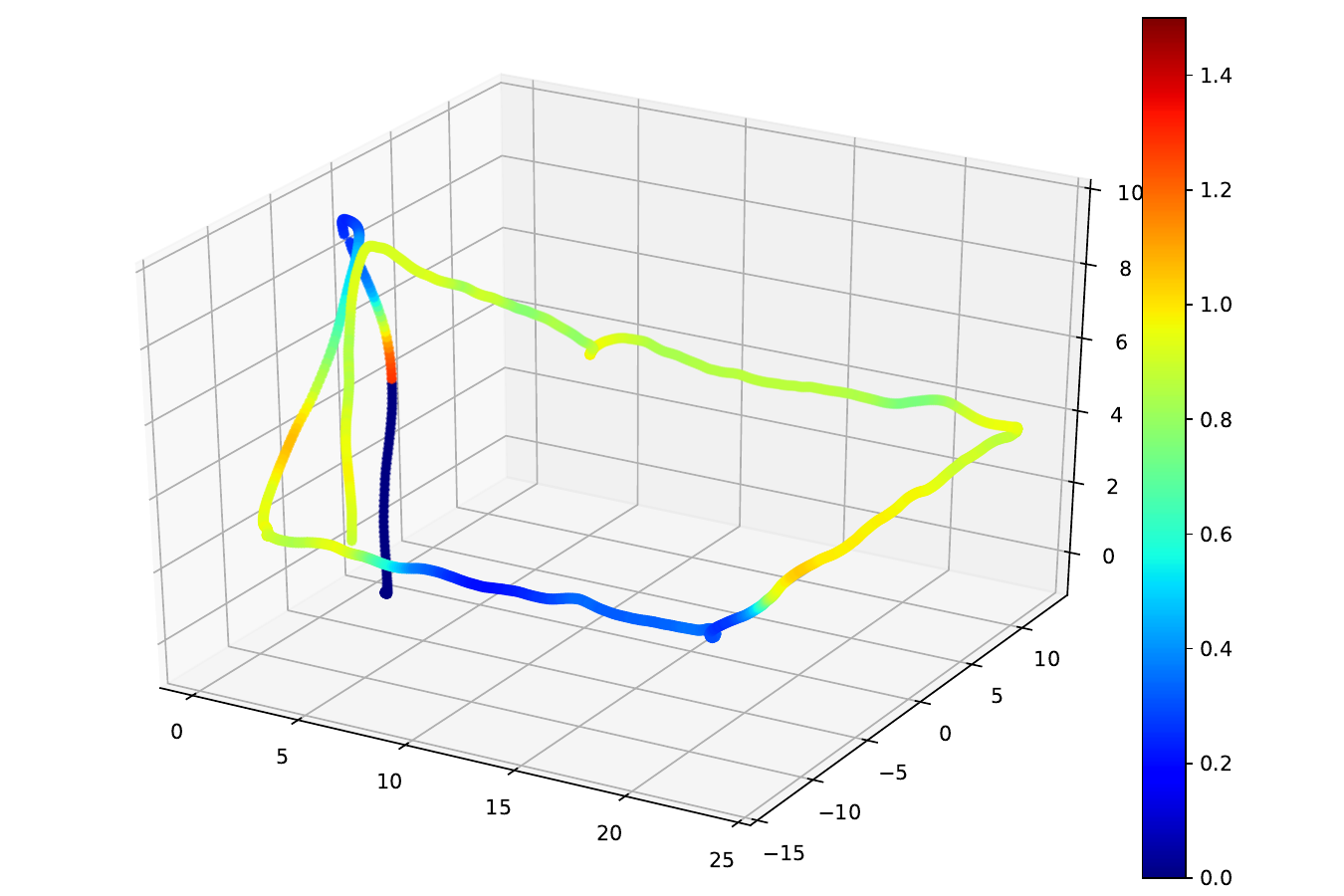}\label{uncertainty2}}\\
        \subfloat[Label 3 prediction]{\centering\includegraphics[width=0.4\linewidth]{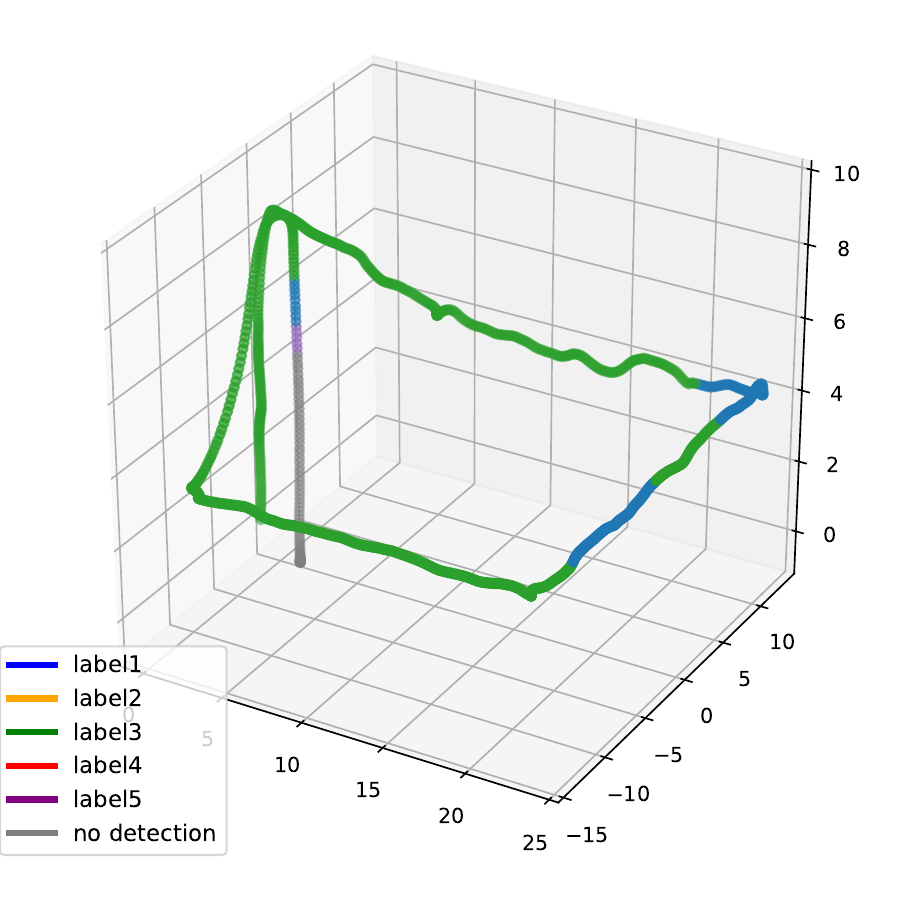}\label{acc3}}
       	\subfloat[Label 3 uncertainty]{\centering\includegraphics[width=0.55\linewidth]{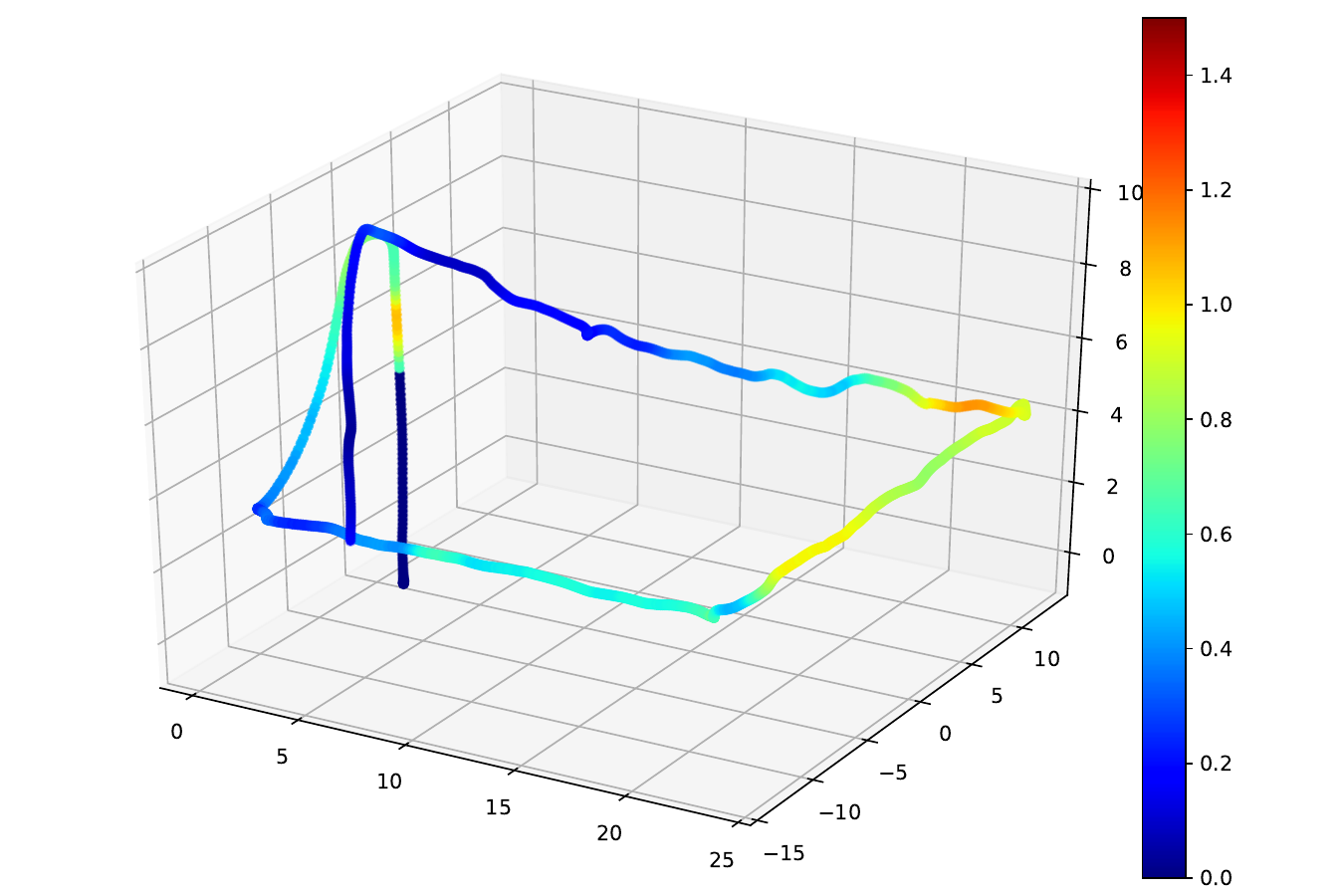}\label{uncertainty3}}\\
        \subfloat[Label 4 prediction]{\centering\includegraphics[width=0.4\linewidth]{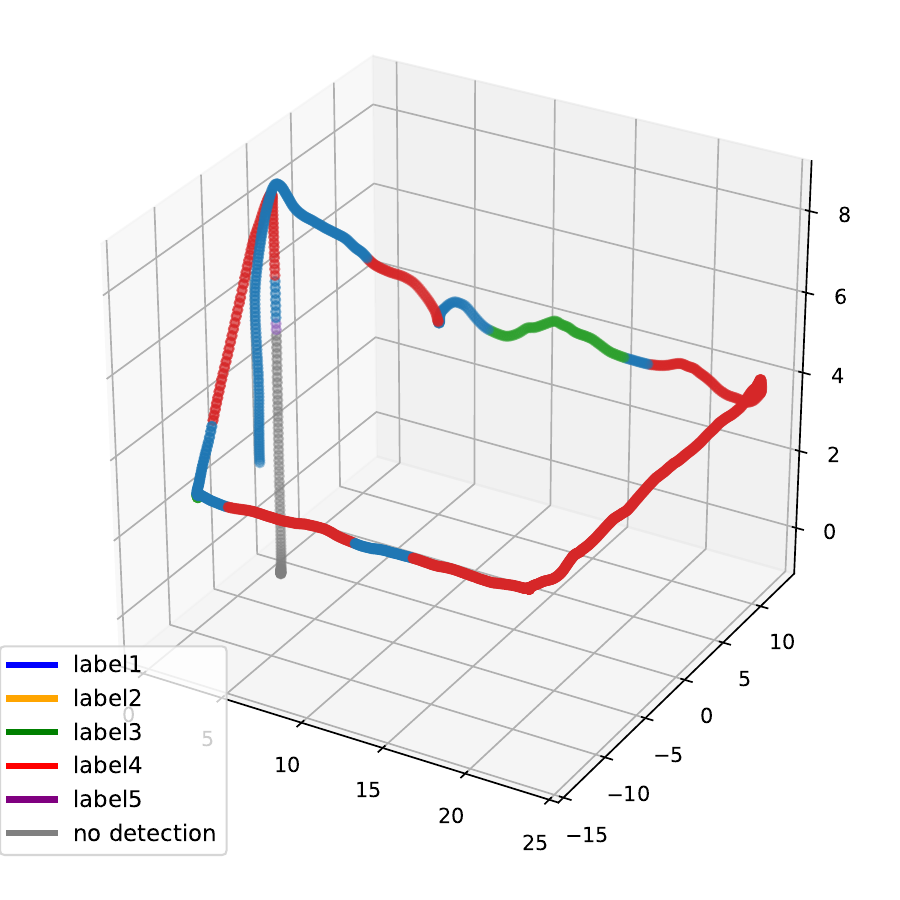}\label{acc4}}
       	\subfloat[Label 4 uncertainty]{\centering\includegraphics[width=0.55\linewidth]{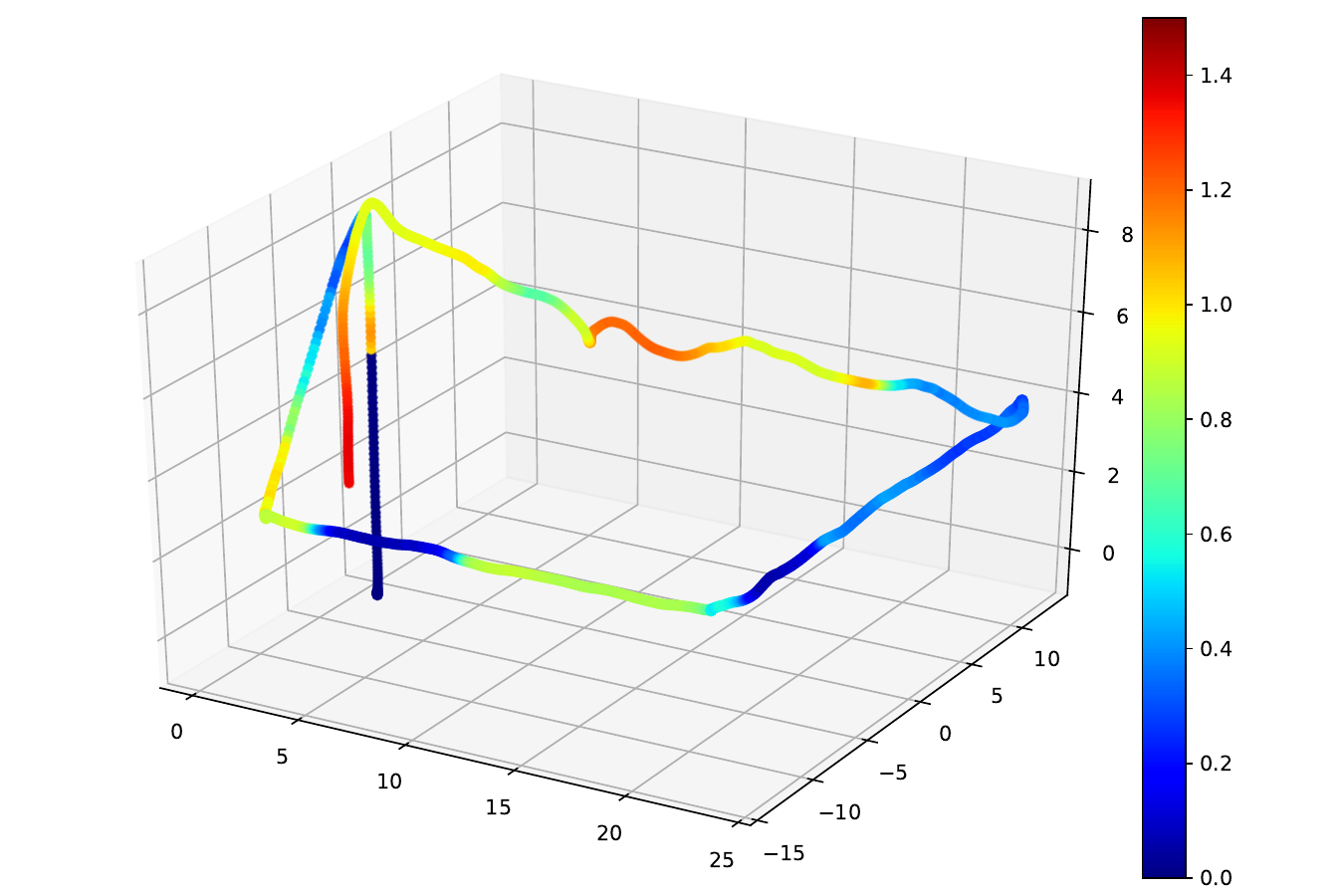}\label{uncertainty4}}\\
        \subfloat[Label 5 prediction]{\centering\includegraphics[width=0.4\linewidth]{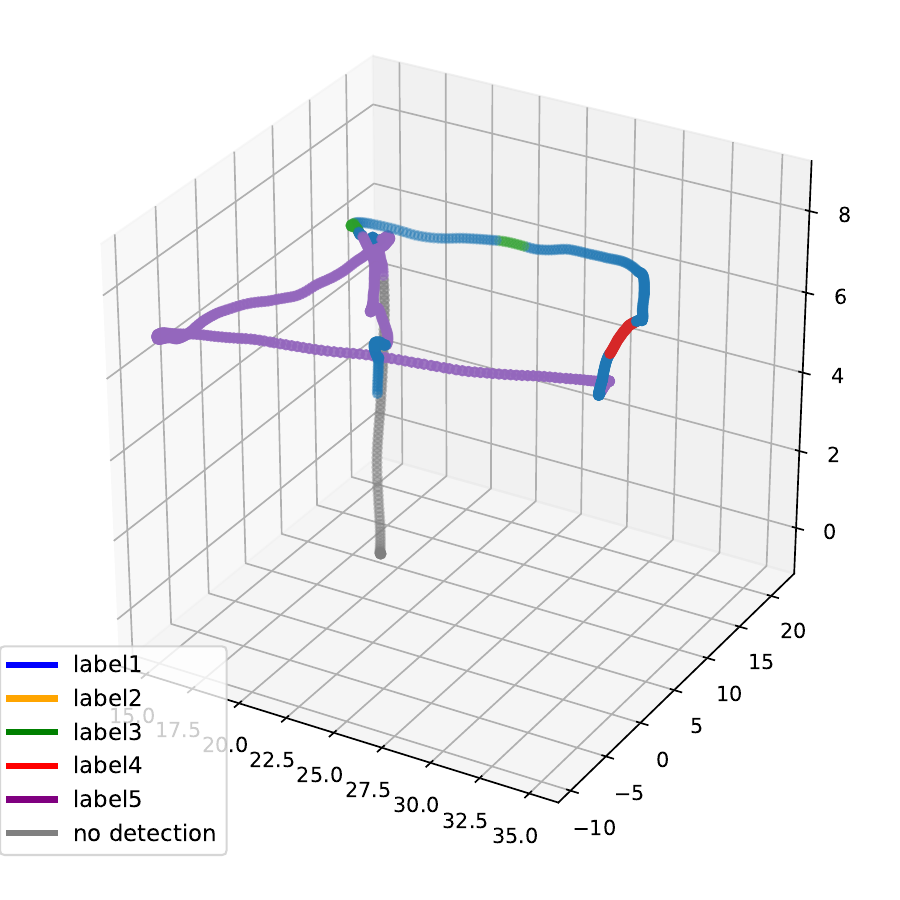}\label{acc5}}
       	\subfloat[Label 5 uncertainty]{\centering\includegraphics[width=0.55\linewidth]{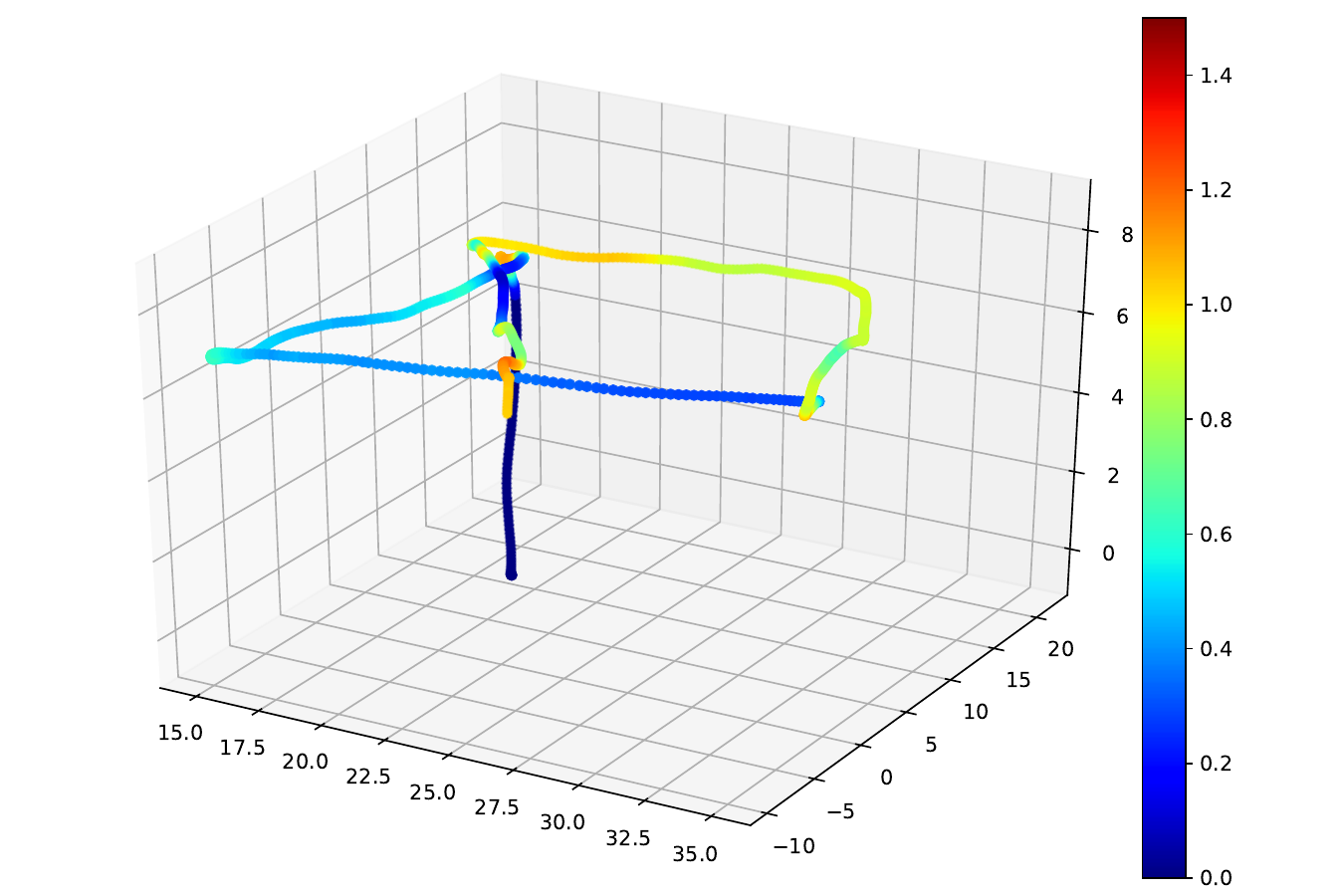}\label{uncertainty5}}\\
	\caption{The prediction result and uncertainty value at each time step for the five fault categories in outdoor real-flight}
	\label{traj_uncertainty}
\end{figure}

In order to gain a deeper understanding of both the EDDCNN model and the UFC decision-making process, we analyzed the prediction results and uncertainty values at each time step of the flying trajectories, as shown in Fig. \ref{traj_uncertainty}. As shown, the color coding in prediction figures (on the left) indicates the predicted labels of the model at the current time step. The color coding in the uncertainty figures (on the right) indicates the degree of confidence that the model has in its predictions, with blue representing high confidence and red representing low confidence. Since the time-window length of the input signal was 8 seconds, the fault diagnosis process began 8 seconds after the UAV took off. As depicted in Fig. \ref{uncertainty1}, the uncertainty values for the entire flight trajectory were high, leading to the filtering out of prediction results by the UFC. Moreover, as shown in Fig. \ref{acc2}-Fig. \ref{uncertainty5}, all incorrect predictions were associated with high predictive uncertainty.
The prediction of UFC became more accurate when it was confident in its predictions. This observation clarified why our UFC can enhance the accuracy of the EDDCNN model.

\section{Conclusion}\label{Conclusion}
In this paper, we have presented a feasible solution to the challenging problem of sim-to-real UAV fault diagnosis in a windy environment. By leveraging our proposed EDDCNN model and UFC, we achieved a substantial increase in the average accuracy of the classifier, from 58.3\% to 100\%. Despite predictions with high uncertainty being filtered out by UFC, the average data usage rate for the four fault categories can reach 33.6\%, which is acceptable for real flights. The visualization of the prediction and uncertainty values made it easier to comprehend why our UFC can improve the classification accuracy in outdoor real-flight data. In the future, we intend to further enhance the data usage rate of our UFC.

\bibliographystyle{IEEEtran}
\bibliography{IEEEabrv,BIB_xx-TIE-xxxx}\ 

\end{document}